%% file: neurips_2026.tex
\documentclass{article}
\usepackage{algorithm}
\usepackage{algorithmic}

\usepackage[main, final]{neurips_2026}

\usepackage[utf8]{inputenc} 
\usepackage[T1]{fontenc}    
\usepackage{hyperref}       
\usepackage{url}            
\usepackage{booktabs}       
\usepackage{amsfonts}       
\usepackage{nicefrac}       
\usepackage{microtype}      
\usepackage{xcolor}         

\usepackage{multirow}
\usepackage{colortbl}
\usepackage{amsmath}
\usepackage{amssymb}
\usepackage{wrapfig}
\usepackage{enumitem}
\usepackage{amsthm}
\usepackage{graphicx}
\usepackage{array}
\usepackage{tabularx}
\usepackage{framed}
\usepackage{subcaption}
\usepackage{float}

\definecolor{tagblue}{RGB}{30,100,200}
\definecolor{tagorange}{RGB}{220,120,30}
\definecolor{tagred}{RGB}{200,40,40}
\definecolor{tagteal}{RGB}{20,140,130}
\definecolor{bgmint}{RGB}{220,240,225}
\definecolor{MMGraph}{HTML}{6A0DAD}
\definecolor{gaingreen}{HTML}{2E7D32}
\definecolor{promptbg}{RGB}{246,246,250}
\colorlet{shadecolor}{promptbg}
\providecommand{\stat}[2]{#1\,{\fontsize{6pt}{7pt}\selectfont $\pm$#2}}
\providecommand{\statb}[2]{\textbf{#1}\,{\fontsize{6pt}{7pt}\selectfont $\pm$#2}}
\newcommand{\tg}[2]{\textcolor{#1}{\texttt{<#2>}\ldots\texttt{</#2>}}}

\title{MMGraph-Agent: Agentic Multimodal RAG via Cache-Inspired Multimodal Knowledge HyperGraphs}

\vspace{-5.5mm}

\author{
{\bf Haoran Luo\textsuperscript{\rm 1}\textsuperscript{$\dagger$}, Ziyue Zhu\textsuperscript{\rm 2}\textsuperscript{$*$}, Shangyang Wu\textsuperscript{\rm 2}, Linhao Luo\textsuperscript{\rm 3}, Yikai Guo\textsuperscript{\rm 4}, Qika Lin\textsuperscript{\rm 5},} \\
{\bf Fangzhi Xu\textsuperscript{\rm 6}, Jiapu Wang\textsuperscript{\rm 8}, Xiaobao Wu\textsuperscript{\rm 9}, Yifan Zhu\textsuperscript{\rm 2}, Luu Anh Tuan\textsuperscript{\rm 1,7}\textsuperscript{$*$}} \\
         \textsuperscript{1}Nanyang Technological University\ \ \textsuperscript{2}Beijing University of Posts and Telecommunications \\
         \textsuperscript{3}Monash University\ \ \textsuperscript{4}Beijing Institute of Computer Technology and Application \\ 
         \textsuperscript{5}National University of Singapore\ \ \textsuperscript{6}Xi'an Jiaotong University\ \ \textsuperscript{7}VinUniversity\\ 
         \textsuperscript{8}Nanjing University of Science and Technology\ \ \textsuperscript{9}Shanghai Jiao Tong University\\ 
         \texttt{haoran.luo@ntu.edu.sg, zyzhu@bupt.edu.cn, anhtuan.luu@ntu.edu.sg}
}

\begin{document}

\maketitle
{\renewcommand{\thefootnote}{$\dagger$}\footnotetext{Project Lead.\quad\textsuperscript{$*$}Corresponding Authors.\quad\textsuperscript{1}\url{https://github.com/LHRLAB/MMGraph-Agent}}}

\vspace{-4.5mm}

\begin{abstract}
Driven by increasingly complex real-world applications, Retrieval-Augmented Generation (RAG) has evolved from textual pipelines to multimodal settings, comprising two complementary dimensions: knowledge organization and retrieval enhancement. However, existing methods still face challenges of high construction cost, context window limitations, and the lack of unified training across offline and online knowledge sources. To address these challenges, we propose \textbf{MMGraph-Agent}, a unified multimodal RAG framework based on cache-inspired multimodal knowledge hypergraphs. MMGraph-Agent enables extremely lightweight hypergraph construction with zero API cost and dynamic memory-based retrieval that alleviates long-context issues. Experiments across offline, online, and joint retrieval settings demonstrate consistent improvements in performance, efficiency, and architectural unification. Our software and data are publicly available\textsuperscript{1}.
\end{abstract}

\vspace{-3.7mm}

\section{Introduction}\label{sec:intro}

\label{Section1}

\begin{wrapfigure}{r}{0.51\textwidth}
\vspace{-5.5mm}
\centering
\includegraphics[width=0.98\linewidth]{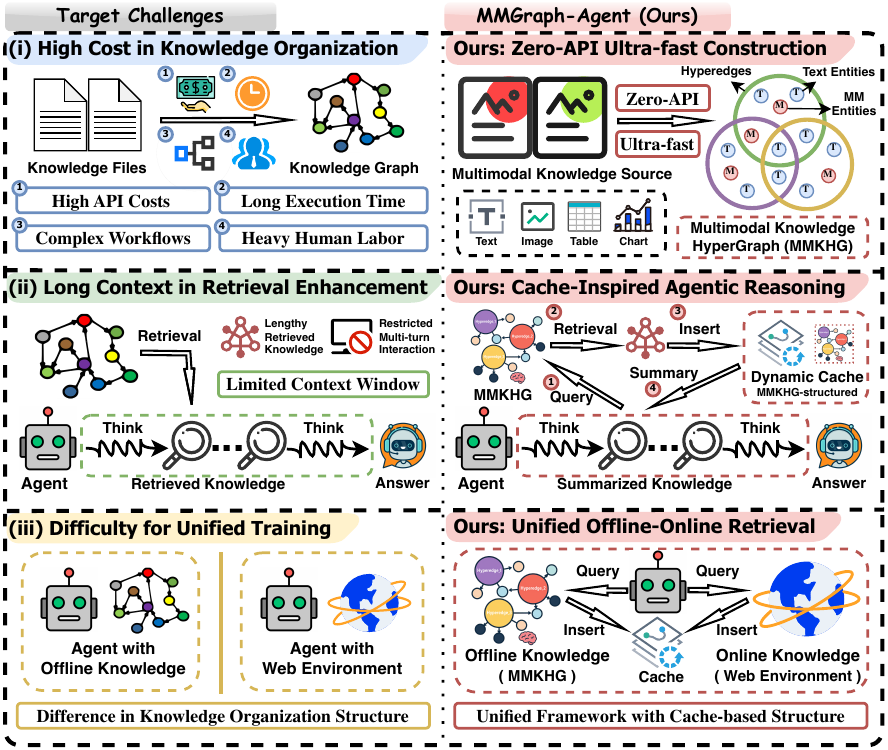}
\caption{Target challenges and MMGraph-Agent.}
\label{F1}
\vspace{-4mm}
\end{wrapfigure}

Retrieval-Augmented Generation (RAG)~\citep{RAG} systems have evolved from text-only pipelines to multimodal settings~\citep{VisRAG,VisRAG2.0} spanning text, images, tables, and diagrams. Building on this paradigm, multimodal RAG thus decomposes into two dimensions: \textbf{(i) Knowledge Organization}, which structures multimodal knowledge into coherent representations~\citep{GraphRAG, HyperGraphRAG, RAG-Anything}, and \textbf{(ii) Retrieval Enhancement}, which governs how relevant knowledge is retrieved and then injected into the generation process~\citep{Search-R1,MMSearch-R1}.

Existing RAG approaches range from chunk-based text retrieval to their image-level extensions~\citep{VisRAG,VisRAG2.0}. As for knowledge organization, GraphRAG~\citep{GraphRAG} and HyperGraphRAG~\citep{HyperGraphRAG} use entity-relation graphs to enhance knowledge structures, whereas MMGraphRAG~\citep{MMGraphRAG} and RAG-Anything~\citep{RAG-Anything} extend this to multimodal knowledge graphs and add multimodal entities. For retrieval enhancement, Search-R1~\citep{Search-R1} and Graph-R1~\citep{Graph-R1} employ agentic reinforcement learning for active retrieval, and MMSearch-R1~\citep{MMSearch-R1} further enables multimodal agentic search on the web.

\begin{figure*}[t]
\centering
\includegraphics[width=\linewidth]{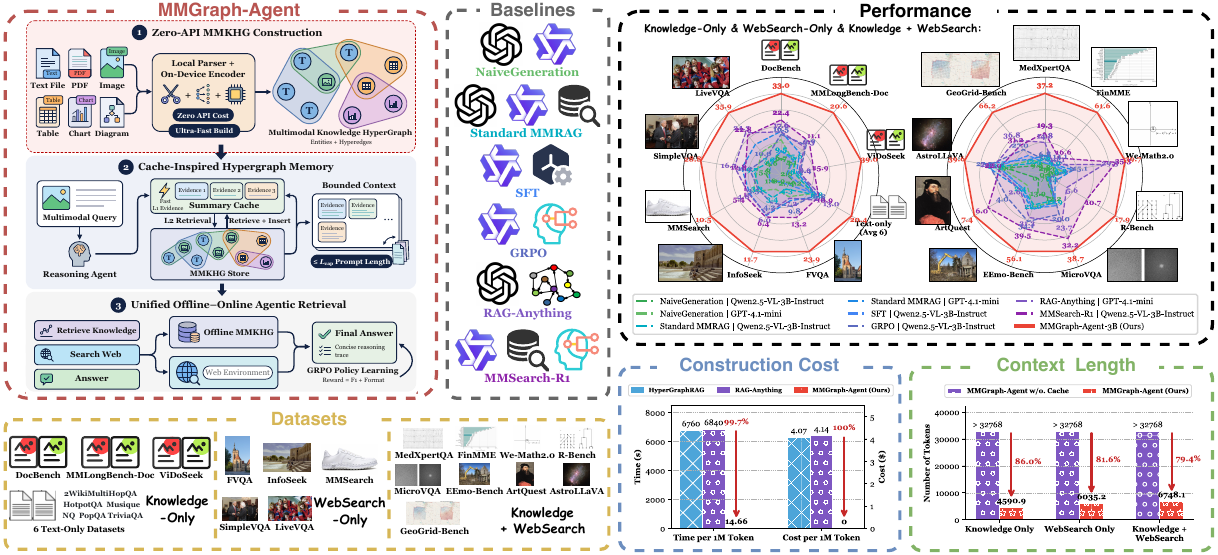}
\caption{Overview of MMGraph-Agent: a multimodal RAG framework built on cache-inspired knowledge hypergraphs supporting unified joint retrieval with reduced cost and context overhead.}
\label{F2}
\end{figure*}

However, as shown in Figure~\ref{F1}, three main challenges remain. \textbf{(i) High Cost in Knowledge Organization.} Constructing multimodal knowledge graphs often incurs substantial costs~\citep{HyperGraphRAG, RAG-Anything}, relying on manual annotation or expensive API calls, which limits scalability. \textbf{(ii) Long Context in Retrieval Enhancement.} Multi-round retrieval is constrained by model context windows~\citep{LongBench}; as retrieved evidence accumulates, context overflow becomes increasingly likely~\citep{Graph-R1,MemSearcher}, degrading reasoning quality. \textbf{(iii) Lack of a Unified Framework.} Unifying local private knowledge~\citep{RAG-Anything} and online web knowledge~\citep{MMSearch-R1} within a single framework is challenging due to structural differences.

To address these challenges, we propose \textbf{MMGraph-Agent}, a unified multimodal RAG framework based on cache-inspired multimodal knowledge hypergraphs. As shown in Figure~\ref{F2}, we first propose a lightweight \textbf{multimodal hypergraph construction method with zero API cost}, substantially reducing both construction cost and build time while still preserving knowledge completeness. We further design a \textbf{cache-inspired multimodal hypergraph memory} that dynamically stores, updates, and serves retrieved multimodal evidence during iterative agentic reasoning, thereby alleviating context window limitations and improving multi-round retrieval efficiency. Furthermore, MMGraph-Agent also integrates multimodal reinforcement learning (RL) to enable \textbf{coordinated retrieval from offline knowledge corpora and online web environments} within a single unified RAG framework.

To validate its effectiveness, we conduct experiments in three settings: Knowledge-Only, WebSearch-Only, and Knowledge+WebSearch. Results demonstrate that MMGraph-Agent consistently outperforms a range of baselines spanning generation, RAG-based, and training-based methods in answer accuracy and retrieval efficiency, while substantially reducing construction cost and context token usage, thereby demonstrating its strong potential for real-world multimodal question answering.

\section{Related Work}

\textbf{Multimodal RAG.}
RAG grounds LLMs with external knowledge~\citep{RAG} and has evolved from text-only pipelines to multimodal settings~\citep{VisRAG,VisRAG2.0}. For \emph{knowledge organization}, prior work moves from flat or hierarchical chunk retrieval~\citep{RAPTOR,Self-RAG} to entity-relation graphs and higher-order hyperedges~\citep{GraphRAG,MedGraphRAG,KAG,LightRAG,HippoRAG2,HopRAG,GFM-RAG,HyperGraphRAG,PropRAG,CausalRAG,HyperRAG,G-reasoner,Cog-RAG}, further extended to multimodal knowledge graphs and hybrid vector-graph designs~\citep{MMGraphRAG,RAG-Anything,M3KG-RAG,MegaRAG,MG2-RAG,mKG-RAG,EvoGraph-R1}. In this work, MMGraph-Agent builds a multimodal hypergraph that preserves cross-modal relational structure at zero API cost and high build speed.

\textbf{Agentic Retrieval.}
For \emph{retrieval enhancement}, LLM-based agents reframe retrieval as sequential decision-making, letting the model decide \emph{when} and \emph{what} to retrieve~\citep{ReAct,AgentSurvey}. Recent methods drive iterative, reasoning-grounded retrieval over textual and graph corpora~\citep{Search-R1,R1-Searcher,Graph-R1} and extend it to multimodal and open-web search~\citep{MMSearch-R1,DeepEyesV2,DeepImageSearch,ZeroSearch,WebDancer,WebSailor,DeepDive,GraphSearch,FlowSearch,TonyiDeepResearch}. Building on this line of work, MMGraph-Agent introduces a cache-inspired hypergraph memory to keep context bounded across long retrieval trajectories, and a unified policy that jointly retrieves from offline and online sources.

\clearpage

\begin{table}[t]
\centering
\caption{Comparison of representative baselines. StandardRAG/MMRAG: flat (multi-)modal vector RAG; HyperGraphRAG: $n$-ary entity hyperedges; RAG-Anything: paired structural and textual graphs; Graph-R1/MMSearch-R1: RL agents over a hypergraph or web; MMGraph-Agent (Ours): unifies a multimodal hypergraph with live web data via hierarchical cache/graph/web retrieval.}
\label{tab:comparison}
\vspace{6pt}
\fontsize{10.5pt}{13pt}\selectfont
\centering
\resizebox{\linewidth}{!}{
\begin{tabular}{p{3.3cm}|p{5.7cm}|p{6.3cm}}
\toprule
\textbf{Method} & \textbf{Knowledge Organization} & \textbf{Retrieval Enhancement} \\
\midrule
StandardRAG~\citep{RAG}             & $\mathcal{K}=\{c_i\}_{i=1}^{N},\ \mathbf{e}_i=\mathrm{Enc}(c_i)$ & $\mathcal{R}(q)=\mathrm{Top\text{-}}k_{c_i\in\mathcal{K}}\,\mathrm{sim}(\mathrm{Enc}(q),\mathbf{e}_i)$ \\
StandardMMRAG~\citep{VisRAG}        & $\mathcal{K}=\{I_i\}_{i=1}^{N},\ \mathbf{e}_i=\mathrm{MMEnc}(I_i)$ & $\mathcal{R}(q)=\mathrm{Top\text{-}}k_{I_i\in\mathcal{K}}\,\mathrm{sim}(\mathrm{MMEnc}(q),\mathbf{e}_i)$ \\
\midrule
HyperGraphRAG~\citep{HyperGraphRAG} & $\mathcal{G}_H=(\mathcal{V},\mathcal{E}_H),\ e\in 2^{\mathcal{V}}$\newline $n$-ary relational entity hyperedges. & $\mathcal{V}_q\to\mathcal{E}_q=\{e:\,e\cap\mathcal{V}_q\neq\emptyset\}$\newline Staged entity-then-hyperedge retrieval. \\
RAG-Anything~\citep{RAG-Anything}   & $\mathcal{G}=(\mathcal{G}_s,\mathcal{G}_t)$\newline Dual structural and textual graphs. & $\mathcal{R}(q)=\mathrm{Nav}(\mathcal{G}_s,q)\cup\mathrm{Sim}(\mathcal{G}_t,q)$\newline Graph navigation with semantic matching. \\
\midrule
Graph-R1~\citep{Graph-R1}           & $\mathcal{G}_H=(\mathcal{V},\mathcal{E}_H)$\newline Lightweight higher-order hypergraph. & $a_t\sim\pi_\theta(\cdot\mid s_t)$\newline RL-driven multi-turn hyperedge search. \\
MMSearch-R1~\citep{MMSearch-R1}     & $\mathcal{W},\ \mathcal{T}=\{\mathrm{text},\mathrm{img}\}$\newline Dynamic multimodal text-image web. & $a_t\sim\pi_\theta(\cdot\mid s_t,\mathcal{T})$\newline RL-driven multi-turn web search agent. \\
\midrule
\textbf{MMGraph-Agent (Ours)}       & $\mathcal{G}_H^m\cup\mathcal{W}$\newline Multimodal hypergraph with web data. & $a_t\sim\pi_\theta(\cdot\mid s_t,\mathcal{M}_1,\mathcal{G}_H^m,\mathcal{W})$\newline Hierarchical cache/graph/web retrieval. \\
\bottomrule
\end{tabular}
}
\end{table}

\section{Methodology: MMGraph-Agent}
\label{sec:method}

\noindent\textbf{Problem Formulation.} We study multimodal retrieval-augmented question answering over a static private corpus and the live web; Table~\ref{tab:comparison} contrasts how representative paradigms instantiate knowledge organization and retrieval enhancement. Formally, let $\mathcal{K}\!=\!\{d_i\}_{i=1}^{N}$ be a static corpus of $N$ multimodal documents over modalities $\mathbb{M}\!=\!\{\text{text},\text{image},\text{table},\text{diagram}\}$, and $\mathcal{W}$ a dynamic web corpus reachable only through a search interface $\mathcal{S}$. Given a query $(q,v_q)$ with text $q$ and an optional image $v_q$, the task is to learn a policy $\pi_{\theta}$ that, over a horizon $T$, emits a trajectory $\tau\!=\!(a_0,\ldots,a_T)$ scheduling offline retrieval $\mathcal{R}_{\mathrm{off}}$ on an indexed view $\phi(\mathcal{K})$, online retrieval $\mathcal{R}_{\mathrm{on}}$ on $\mathcal{W}$ through $\mathcal{S}$, and the final answer $y_{\tau}$, jointly optimized by
\begin{equation}
\tau\sim\pi_{\theta}\bigl(\cdot\mid q,v_q;\phi(\mathcal{K}),\mathcal{S}(\mathcal{W})\bigr),\qquad
\mathcal{J}(\pi_{\theta})=\mathbb{E}_{(q,v_q,y^*)\sim\mathcal{D}}\bigl[R_{F_{1}}(y_{\tau},y^*)\bigr],
\end{equation}
where $\phi\!:\!\mathcal{K}\!\to\!\mathcal{Z}$ is the offline organizer producing the indexed knowledge store $\mathcal{Z}$, $\mathcal{D}$ the training distribution, $y^{*}$ the gold answer, and $R_{F_{1}}$ the token-level F$_{1}$ reward, maximized via the RL objective of \S\ref{sec:rl}. The three modules below address the cost, context, and unification challenges in turn.

\subsection{Zero-API-Cost Multimodal Hypergraph Construction}
\label{sec:construction}

\noindent We instantiate $\phi$ at \emph{zero API cost}: our local parser--extractor replaces the paid LLM extractors~\citep{HyperGraphRAG,RAG-Anything,MMGraphRAG} while achieving modality-inclusive construction and within-unit higher-order relations.

\noindent\textbf{Modality-Aware Parsing and Local Entity Extraction.} Multimodal documents in $\mathcal{K}$ interleave paragraphs, figures, tables and diagrams whose semantic units rarely align with token boundaries, so we realize $\phi$ as a parser--extractor pipeline that runs entirely on local components and never queries any paid LLM. A rule-based parser $\psi$ first slices every document $d\!\in\!\mathcal{K}$ into modality-tagged units, i.e., spaCy-segmented sentences as textual units and attached figures as image units (tables and diagrams enter as flattened text or rendered images); a local extractor $\pi_{\mathrm{ext}}$ then applies a spaCy named-entity recognizer (NER) to every textual unit, while a frozen multimodal encoder $f\!:\!\mathcal{X}\!\to\!\mathbb{R}^{D}$ embeds the extracted entities and the image units into a shared space in one streaming pass,
\begin{equation}
\psi(d)\!=\!\bigl\{u_{j}^{\mu_{j}}\bigr\}_{j=1}^{J(d)},\quad
\mathcal{V}^{m}\!=\!\!\!\bigcup_{d\in\mathcal{K}}\!\bigcup_{u\in\psi_{\mathrm{txt}}(d)}\!\!\!\pi_{\mathrm{ext}}(u),\quad
\mathbf{E}_{\mathcal{V}}\!=\!\bigl\{f(v):v\!\in\!\mathcal{V}^{m}\bigr\}\!\subset\!\mathbb{R}^{D},
\end{equation}
where $u_{j}^{\mu_{j}}$ is the $j$-th unit tagged by modality $\mu_{j}\!\in\!\mathbb{M}$, $\psi_{\mathrm{txt}}(d)\!\subseteq\!\psi(d)$ its textual units, $J(d)$ the unit count of $d$, $\mathcal{V}^{m}$ the entity set extracted from textual units (the superscript $m$ marks multimodal-corpus constructs), $\mathcal{X}$ the joint space of entities, hyperedges, images and queries shared by $f$, and $\mathbf{E}_{\mathcal{V}}$ the $D$-dimensional entity embedding bank; image units, collected as $\mathcal{I}\!=\!\{u:\mu(u)\!=\!\mathrm{image}\}$ with $\mu(u)$ the modality tag of unit $u$, bypass $\pi_{\mathrm{ext}}$ and are embedded directly by $f$ into a parallel visual bank $\mathbf{E}_{\mathcal{I}}\!=\!\{f(u):u\!\in\!\mathcal{I}\}$, sharing one retrieval space with textual items. As neither $\psi$ nor $\pi_{\mathrm{ext}}$ calls a paid LLM, the entity layer of $\phi$ is built at \emph{zero API cost} and re-runs locally on new documents.

\noindent\textbf{Hyperedge Induction and Streaming Index.} Pairwise edges collapse the higher-order context that grounds multimodal reasoning, so all entities co-occurring inside the same parsed sentence form an $n$-ary hyperedge $e_{H}$ that carries the full sentence as fragment payload, and these hyperedges are embedded by the same encoder $f$ to populate a streaming index alongside the entity and visual banks,
\begin{equation}
\mathcal{E}_{H}\!=\!\bigl\{(\pi_{\mathrm{ext}}(u),u):u\!\in\!\psi_{\mathrm{txt}}(d),\,d\!\in\!\mathcal{K},\,|\pi_{\mathrm{ext}}(u)|\!\geq\!1\bigr\},\;\,
\mathbf{E}_{\mathcal{E}_{H}}\!=\!\{f(e_{H})\}_{e_{H}\in\mathcal{E}_{H}},\;\,
\mathcal{G}_{H}^{m}\!=\!(\mathcal{V}^{m},\mathcal{E}_{H}),
\end{equation}
where each $e_{H}$ has arity $n\!=\!|\pi_{\mathrm{ext}}(u)|\!\geq\!1$ and retains its host sentence $u$ as payload, $\mathbf{E}_{\mathcal{E}_{H}}\!\subset\!\mathbb{R}^{D}$ is the hyperedge bank, and $\mathcal{G}_{H}^{m}$ together with the visual bank $\mathbf{E}_{\mathcal{I}}$ constitutes the multimodal hypergraph store. Cross-modal relevance then reduces to the inner product $\mathrm{sim}(x,x')\!=\!\langle f(x),f(x')\rangle$ on $\mathcal{X}$, used uniformly by every downstream retrieval call: $f$ is a frozen CLIP backbone with $D\!=\!768$ and $\mathcal{G}_{H}^{m}$ is backed by flat FAISS inner-product indices, giving fast append-only insertion and single-call \emph{exact} Top-$k$ retrieval (for any requested depth $k$; \S\ref{sec:memory} uses $k\!=\!C$) across all the modalities of the corpus.

\textbf{Proposition 1.} \textit{$\mathcal{G}_{H}^{m}$ attains entity coverage comparable to an LLM extractor at zero API cost.}
\vspace{-3mm}
\begin{proof}
We provide experimental results in \S\ref{sec:rq3} and formal analyses in Appendix~\ref{app:proof1}.
\end{proof}

\begin{figure*}[t]
\centering
\includegraphics[width=\linewidth]{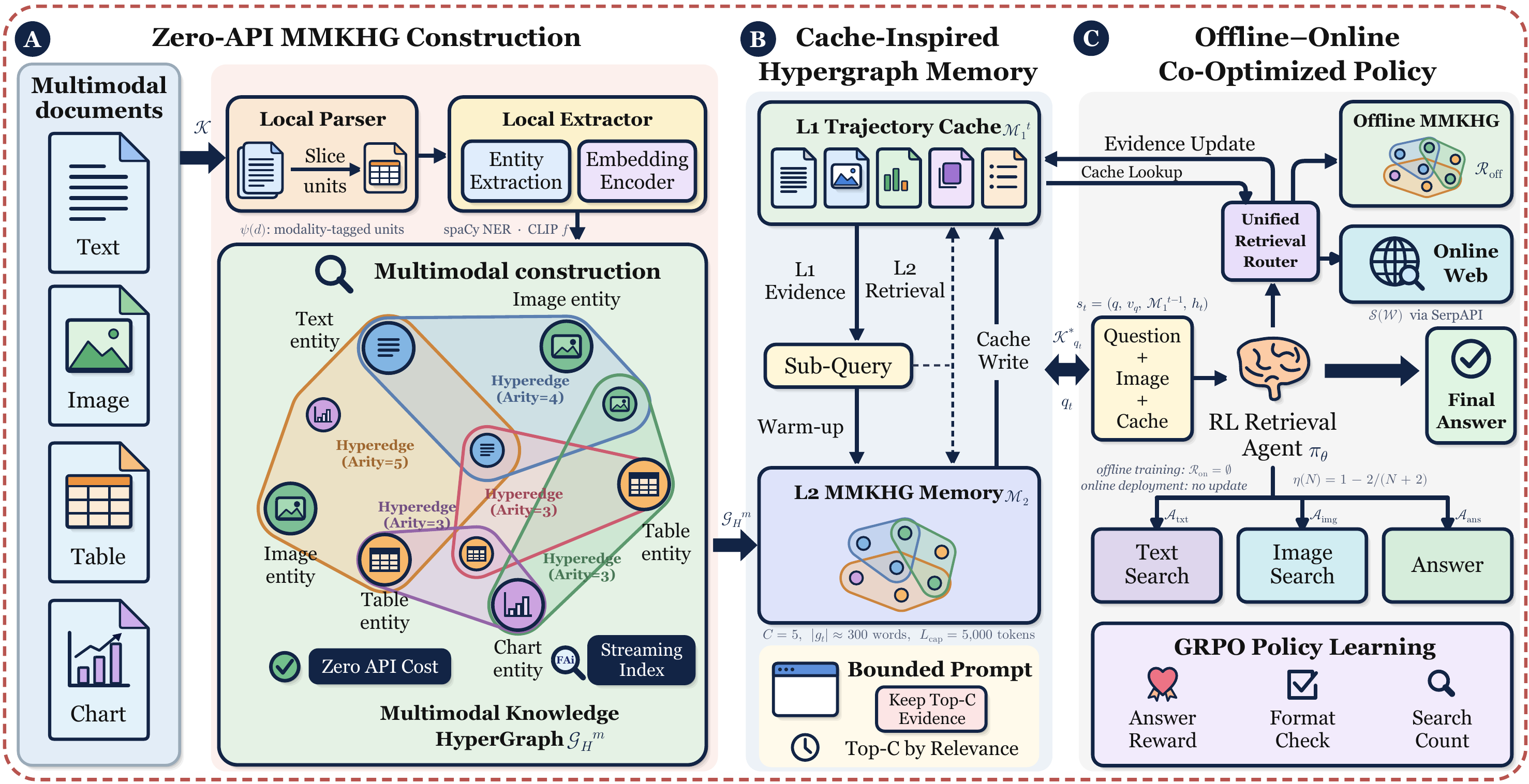}
\caption{Methodology of MMGraph-Agent: zero-API multimodal hypergraph (MMKHG) wrapped by a two-tier cache memory and queried by an offline-trained, online-deployed retrieval policy.}
\label{F3}
\end{figure*}

\subsection{Cache-Inspired Multimodal Hypergraph Memory}
\label{sec:memory}

\noindent On top of $\mathcal{G}_{H}^{m}$, an L1/L2 cache memory dynamically stores and reuses retrieved evidence~\citep{Graph-R1,LongBench,HGMem,MAGE} so the prompt stays bounded uniformly across the multi-round retrieval horizon.

\noindent\textbf{Two-Tier Memory with Per-Round Warm-Up.} Multi-round retrieval inflates the prompt and risks context overflow, so we mirror the CPU L1/L2 cache hierarchy and split $\mathcal{R}_{\mathrm{off}}$ across two tiers: the L2 backend $\mathcal{M}_{2}$ persistently stores the full hypergraph store of \S\ref{sec:construction}, while the L1 per-trajectory cache $\mathcal{M}_{1}^{t}$ accumulates the evidence retrieved so far. At round $t$, given the agent state $s_{t}\!=\!(q,v_{q},\mathcal{M}_{1}^{t-1},h_{t})$ with reasoning trace $h_{t}$, the policy emits the sub-query $q_{t}\!\in\!\mathcal{X}$ inside its retrieval action (\S\ref{sec:rl}); the controller consults L2 (and, when enabled, the web channel $\mathcal{R}_{\mathrm{on}}$), admits the fresh evidence into L1, and serves the merged Top-$C$ evidence from the unified L1 index,
\begin{equation}
\mathcal{R}_{2}(q_{t})\!=\!\operatornamewithlimits{Top\text{-}C}_{x\in\mathcal{M}_{2}}\!\langle f(q_{t}),f(x)\rangle,\;\;
\mathcal{M}_{1}^{t}\!=\!\mathcal{M}_{1}^{t-1}\cup\mathcal{R}_{2}(q_{t})\cup\mathcal{R}_{\mathrm{on}}(q_{t}),\;\;
\mathcal{K}^{*}_{q_{t}}\!=\!\operatornamewithlimits{Top\text{-}C}_{x\in\mathcal{M}_{1}^{t}}\!\langle f(q_{t}),f(x)\rangle,
\end{equation}
where $C$ is the per-bank retrieval budget shared by both tiers, $\mathcal{R}_{\mathrm{on}}(q_{t})$ the web evidence (empty when disabled), and $\mathcal{K}^{*}_{q_{t}}$ the merged evidence returned to the agent. The empty initial cache $\mathcal{M}_{1}^{0}\!=\!\varnothing$ warms the working set from $\mathcal{G}_{H}^{m}$ rather than policy priors, and later rounds re-serve evidence from L1.

\noindent\textbf{Digest-Bounded Prompt Admission.} The admission rule decides what enters the context: raw concatenation grows the prompt linearly in $T$, so each evidence block is truncated to $L_{\mathrm{max}}$ words, condensed by the policy into a digest $g_{t}$, and the digest is admitted, with the rollout halted at $L_{\mathrm{cap}}$,
\begin{equation}
\mathrm{prompt}(s_{t+1})\!=\!\mathrm{prompt}(s_{t})\oplus g_{t},\;\;
|g_{t}|\!\leq\!L_{\mathrm{dig}},\;\;
|\mathrm{prompt}(s_{t})|\!\leq\!\min\bigl(|\mathrm{prompt}(s_{0})|\!+\!t\!\cdot\!L_{\mathrm{dig}},\,L_{\mathrm{cap}}\bigr),
\end{equation}
where $\oplus$ appends one conversation turn, $g_{t}$ summarizes $\mathcal{K}^{*}_{q_{t}}$, and $|\cdot|$ measures length in tokens ($L_{\mathrm{max}}$ is counted in words; digests fit the $2{,}048$-token turn limit). We set $C\!=\!5$, $L_{\mathrm{max}}\!=\!1{,}200$ words, a ${\approx}300$-word digest, $L_{\mathrm{cap}}\!=\!5{,}000$ tokens; the prompt thus stays bounded uniformly in $T$ while every retrieved item remains reachable in L1 and the full $\mathcal{G}_{H}^{m}$ in L2.

\textbf{Proposition 2.} \textit{The two-tier memory bounds the prompt uniformly over $T$ with $\mathcal{G}_{H}^{m}$ fully reachable.}
\vspace{-3mm}
\begin{proof}
We provide experimental results in \S\ref{sec:rq4} and theoretical proofs in Appendix~\ref{app:proof2}.
\end{proof}

\begin{table}[t]
\centering
\begin{tabular}{p{13.5cm}}
\toprule
You are a helpful assistant. Answer the given question based on the provided images; you can query the knowledge base as many times as you want. You must first conduct reasoning inside \tg{tagblue}{reason}. For image search, append \textcolor{tagorange}{\texttt{<search><img></search>}} behind the reasoning, which runs Google Lens on the original image; for text search, output a concise query inside \tg{tagorange}{text\_search}, answered by KG/Memory retrieval (and Google Search when enabled). Retrieved evidence returns inside \tg{tagteal}{information}; when you have the final answer, output it inside \tg{tagred}{answer}. Here is the image and the question: \texttt{<image>} \textcolor{tagred}{\textit{question}}. Assistant: \\
\bottomrule
\end{tabular}
\vspace{2.2mm}
\caption{Prompt template used by MMGraph-Agent during rollout. At inference time, \textcolor{tagred}{\texttt{question}} is replaced with the user query and \texttt{\textcolor{tagorange}{<image>}} with the visual input; evidence returned from either tool is inserted inside \tg{tagteal}{information} right after the corresponding tag.} 
\vspace{-3mm}
\label{tab:mmgraph-agent-template}
\end{table}

\subsection{Offline-Trained, Online-Deployed Policy Learning}
\label{sec:rl}

\noindent We fit $\pi_{\theta}$ with multimodal RL to schedule offline and online retrieval in one trajectory: training is offline-only, and the source-agnostic interface deploys the trained policy online without updates.

\noindent\textbf{Unified Action Space and Hit-Aware Reward.} Following Table~\ref{tab:mmgraph-agent-template}, $\pi_{\theta}$ samples $a_{t}\!\sim\!\pi_{\theta}(\cdot\!\mid\!s_{t})$ from text-query $\mathcal{A}_{\mathrm{txt}}$, image-query $\mathcal{A}_{\mathrm{img}}$, and termination $\mathcal{A}_{\mathrm{ans}}$ channels in one autoregressive emission; both retrieval channels are routed through the cache (\S\ref{sec:memory}), so the policy emits a single retrieval token whether the hit comes from $\mathcal{G}_{H}^{m}$ or web search via $\mathcal{S}$. We shape the scalar return with a smooth call-count regulator $\eta(\!\cdot\!)$ that pays for the first retrieval and decays the marginal bonus of extra calls (no explicit per-call cost is imposed), plus a structural penalty on ill-formed or retrieval-free trajectories,
\begin{equation}
\mathcal{A}\!=\!\mathcal{A}_{\mathrm{txt}}\!\cup\!\mathcal{A}_{\mathrm{img}}\!\cup\!\mathcal{A}_{\mathrm{ans}},\quad
\eta(N)\!=\!1\!-\!\tfrac{2}{N+2},\quad
r(\tau)\!=\!\mathbb{1}[N_{c}\!\geq\!1]\,\eta(N_{c})\,R_{F_{1}}\!-\!\lambda\bigl(1\!-\!\mathbb{1}_{\mathrm{fmt}}\!\cdot\!\mathbb{1}[N_{c}\!\geq\!1]\bigr),
\end{equation}
where $\tau$ ends in $\mathcal{A}_{\mathrm{ans}}$, $N_{c}$ counts retrievals, $\eta$ is concave with $\eta(0)\!=\!0,\eta(\infty)\!=\!1$, $R_{F_{1}}$ the token-level F$_{1}$ against $y^{*}$ ($0$ if none), $\mathbb{1}_{\mathrm{fmt}}$ flags schema compliance, and $\lambda\!=\!0.5$; the penalty vanishes only on compliant trajectories with $N_{c}\!\geq\!1$, pushing $\pi_{\theta}$ to consult the cache rather than parametric priors.

\noindent\textbf{Cost-Efficient Offline-Only Training.} As SerpAPI is costly per rollout, we keep $\mathcal{R}_{\mathrm{on}}$ disabled during training and enable $\mathcal{W}$ at inference; the unified cache hides the source, so offline gradients transfer once $\mathcal{R}_{\mathrm{on}}$ is enabled. We optimize a GRPO clipped surrogate~\citep{GRPO} against a frozen $\pi_{\mathrm{ref}}$,
\begin{equation}
\hat{A}_{i}\!=\!\frac{r(\tau_{i})\!-\!\mu_{G}}{\sigma_{G}\!+\!\delta},\;\,
\mathcal{J}_{\mathrm{GRPO}}(\theta)\!=\!\mathbb{E}_{\tau_i}\!\!\left[\sum_{i,t}\!\tfrac{1}{G|\tau_{i}|}\!\min\!\bigl(\rho_{t}\hat{A}_{i},\,\mathrm{clip}(\rho_{t},1\!\pm\!\epsilon)\hat{A}_{i}\bigr)\right]\!-\!\kappa\,\mathrm{KL}(\pi_{\theta}\!\parallel\!\pi_{\mathrm{ref}}),
\end{equation}
with $\tau_{i}\!\sim\!\pi_{\theta_{\mathrm{old}}}$ the $G$ rollouts sampled per query, $(\mu_{G},\sigma_{G})$ their group reward mean and standard deviation, $\delta$ a small stability constant, $\rho_{t}$ the importance ratio, $\epsilon$ the clip range, and $\kappa$ the KL weight. We run 75 steps with $G\!=\!8,\epsilon\!=\!0.2,\kappa\!=\!10^{-3}$ on $8\!\times\!$A100 and activate $\mathcal{R}_{\mathrm{on}}$ at inference.

\textbf{Proposition 3.} \textit{The source-agnostic interface deploys the offline-trained $\pi_{\theta}$ online without updates.}
\vspace{-3mm}
\begin{proof}
We provide experimental results in \S\ref{sec:rq5_analysis} and formal analyses in Appendix~\ref{app:proof3}.
\end{proof}

\clearpage

\begin{table*}[t]
\centering
\caption{Main results on $23$ multimodal QA datasets in Knowledge-Only, WebSearch-Only, and Knowledge+WebSearch. Token-level F$_{1}$ (\%) with std.\ over three runs; \textbf{bold} marks the best per row, $\dagger$ flags out-of-distribution (OOD) rows, and \textcolor{gaingreen}{green} is the gain over the second-best baseline.}
\label{tab:main_results}
\setlength{\tabcolsep}{1.1pt}\renewcommand{\arraystretch}{1.15} 
\resizebox{\textwidth}{!}{
\begin{tabular}{lcccccccc|c}
\toprule
\multirow{2}{*}{\textbf{Datasets}} & \multicolumn{2}{c}{NaiveGen} & \multicolumn{2}{c}{StdMMRAG} & SFT & GRPO & RAG-Any & MMS-R1 & MMGraph-Agent-3B \\
\cmidrule(lr){2-3}\cmidrule(lr){4-5}\cmidrule(lr){6-6}\cmidrule(lr){7-7}\cmidrule(lr){8-8}\cmidrule(lr){9-9}
& Qwen2.5-3B & GPT-4.1 & Qwen2.5-3B & GPT-4.1 & Qwen2.5-3B & Qwen2.5-3B & GPT-4.1 & Qwen2.5-3B & (Ours) \\
\midrule

\rowcolor{blue!5} DocBench & \stat{6.74}{0.3} & \stat{14.43}{0.8} & \stat{9.54}{0.4} & \stat{17.79}{0.3} & \stat{9.88}{0.5} & \stat{16.50}{0.7} & \stat{17.57}{0.9} & \stat{22.39}{1.4} & \statb{33.05}{4.4}\,\textcolor{gaingreen}{(+10.66)} \\

\rowcolor{blue!5} MMLongBench-Doc & \stat{4.14}{0.2} & \stat{4.78}{0.3} & \stat{5.63}{0.2} & \stat{7.33}{1.8} & \stat{4.95}{0.3} & \stat{9.91}{1.1} & \stat{9.29}{1.5} & \stat{11.08}{1.1} & \statb{20.64}{2.3}\,\textcolor{gaingreen}{(+9.56)} \\
\rowcolor{blue!5} ViDoSeek$^{\dagger}$ & \stat{2.64}{0.1} & \stat{6.49}{0.4} & \stat{6.32}{0.5} & \stat{7.58}{0.8} & \stat{3.08}{0.6} & \stat{9.37}{0.6} & \stat{8.58}{0.4} & \stat{15.89}{0.8} & \statb{38.95}{3.5}\,\textcolor{gaingreen}{(+23.06)} \\

\rowcolor{cyan!3.5} 2Wiki & \stat{4.11}{0.3} & \stat{15.89}{2.0} & \stat{12.51}{1.3} & \stat{15.94}{1.5} & \stat{12.42}{0.7} & \stat{14.42}{0.7} & \stat{15.30}{1.1} & \stat{15.36}{1.1} & \statb{17.68}{5.3}\,\textcolor{gaingreen}{(+1.74)} \\

\rowcolor{cyan!3.5} HotpotQA & \stat{7.29}{1.3} & \stat{12.79}{1.2} & \stat{11.40}{1.3} & \stat{13.85}{1.2} & \stat{15.16}{1.0} & \stat{15.51}{0.6} & \stat{13.32}{1.0} & \stat{14.72}{1.3} & \statb{20.70}{4.2}\,\textcolor{gaingreen}{(+5.19)} \\

\rowcolor{cyan!3.5} Musique & \stat{0.45}{0.2} & \stat{0.59}{0.3} & \stat{0.65}{0.3} & \stat{0.66}{1.1} & \stat{0.27}{0.9} & \stat{0.98}{1.0} & \stat{1.22}{0.2} & \stat{0.78}{0.7} & \statb{5.67}{1.3}\,\textcolor{gaingreen}{(+4.45)} \\

\rowcolor{cyan!3.5} NQ & \stat{3.10}{0.2} & \stat{6.82}{1.9} & \stat{9.36}{1.2} & \stat{8.83}{1.0} & \stat{7.67}{1.3} & \stat{7.79}{0.7} & \stat{10.99}{1.1} & \stat{10.90}{0.8} & \statb{24.42}{2.9}\,\textcolor{gaingreen}{(+13.43)} \\
\rowcolor{cyan!3.5} PopQA$^{\dagger}$ & \stat{7.09}{1.2} & \stat{10.90}{1.3} & \stat{14.34}{1.5} & \stat{7.29}{1.6} & \stat{7.23}{1.2} & \stat{9.21}{1.7} & \stat{14.74}{1.2} & \stat{11.58}{1.1} & \statb{16.80}{2.7}\,\textcolor{gaingreen}{(+2.06)} \\
\rowcolor{cyan!3.5} TriviaQA$^{\dagger}$ & \stat{3.11}{0.4} & \stat{8.41}{0.3} & \stat{10.82}{1.1} & \stat{19.08}{1.8} & \stat{10.05}{1.7} & \stat{11.15}{1.3} & \stat{22.30}{0.9} & \stat{12.13}{1.3} & \statb{37.37}{5.5}\,\textcolor{gaingreen}{(+15.07)} \\
\midrule
\rowcolor{orange!3.5} FVQA$^{\dagger}$ & \stat{2.66}{0.4} & \stat{6.73}{0.4} & \stat{4.34}{0.1} & \stat{7.95}{0.5} & \stat{2.53}{0.6} & \stat{9.83}{0.8} & \stat{7.17}{0.3} & \stat{13.23}{0.8} & \statb{23.87}{3.9}\,\textcolor{gaingreen}{(+10.64)} \\
\rowcolor{orange!3.5} InfoSeek$^{\dagger}$ & \stat{0.94}{0.1} & \stat{2.16}{0.3} & \stat{3.42}{0.2} & \stat{6.14}{0.2} & \stat{1.00}{0.2} & \stat{4.20}{0.2} & \stat{6.01}{0.2} & \stat{6.39}{0.5} & \statb{11.66}{3.5}\,\textcolor{gaingreen}{(+5.27)} \\
\rowcolor{orange!3.5} MMSearch$^{\dagger}$ & \stat{1.61}{0.1} & \stat{4.16}{0.4} & \stat{3.31}{0.3} & \stat{4.69}{0.4} & \stat{1.60}{0.3} & \stat{3.46}{0.2} & \stat{4.82}{0.1} & \stat{5.37}{0.2} & \statb{10.51}{2.9}\,\textcolor{gaingreen}{(+5.14)} \\
\rowcolor{orange!3.5} SimpleVQA$^{\dagger}$ & \stat{5.83}{0.1} & \stat{8.58}{0.3} & \stat{9.71}{0.6} & \stat{14.07}{0.7} & \stat{4.34}{0.5} & \stat{11.22}{0.4} & \stat{16.70}{0.4} & \stat{12.77}{1.2} & \statb{28.82}{4.0}\,\textcolor{gaingreen}{(+12.12)} \\
\rowcolor{orange!3.5} LiveVQA$^{\dagger}$ & \stat{6.59}{0.9} & \stat{13.21}{0.3} & \stat{7.63}{0.8} & \stat{17.64}{0.2} & \stat{6.48}{0.6} & \stat{10.01}{0.8} & \stat{24.32}{0.7} & \stat{22.84}{0.6} & \statb{35.89}{2.7}\,\textcolor{gaingreen}{(+11.57)} \\
\midrule

\rowcolor{green!5} MedXpertQA & \stat{2.84}{0.2} & \stat{4.93}{0.1} & \stat{7.32}{0.2} & \stat{8.95}{0.1} & \stat{5.39}{0.4} & \stat{6.81}{0.3} & \stat{14.83}{0.2} & \stat{19.27}{0.8} & \statb{37.17}{4.5}\,\textcolor{gaingreen}{(+17.90)} \\

\rowcolor{green!5} FinMME & \stat{5.27}{0.1} & \stat{8.14}{0.5} & \stat{9.91}{0.9} & \stat{10.18}{0.4} & \stat{9.53}{0.8} & \stat{14.12}{0.9} & \stat{10.08}{0.5} & \stat{16.63}{0.9} & \statb{61.56}{4.1}\,\textcolor{gaingreen}{(+44.93)} \\

\rowcolor{green!5} We-Math2.0 & \stat{3.95}{1.9} & \stat{9.19}{2.1} & \stat{6.79}{0.6} & \stat{14.17}{1.8} & \stat{10.92}{1.8} & \stat{15.11}{1.5} & \stat{29.32}{1.8} & \stat{35.29}{3.2} & \statb{39.74}{4.6}\,\textcolor{gaingreen}{(+4.45)} \\

\rowcolor{green!5} R-Bench & \stat{0.07}{0.0} & \stat{0.61}{0.1} & \stat{1.26}{0.4} & \stat{2.62}{0.1} & \stat{0.76}{0.1} & \stat{1.14}{0.1} & \stat{5.56}{0.1} & \stat{10.66}{0.1} & \statb{17.85}{2.6}\,\textcolor{gaingreen}{(+7.19)} \\

\rowcolor{green!5} MicroVQA & \stat{12.18}{0.9} & \stat{13.30}{0.2} & \stat{12.90}{0.8} & \stat{13.87}{0.7} & \stat{15.56}{0.5} & \stat{19.98}{1.0} & \stat{23.72}{0.7} & \stat{32.20}{0.6} & \statb{38.72}{5.1}\,\textcolor{gaingreen}{(+6.52)} \\

\rowcolor{green!5} EEmo-Bench & \stat{13.87}{0.9} & \stat{23.64}{0.2} & \stat{21.97}{0.2} & \stat{25.48}{0.1} & \stat{24.46}{0.2} & \stat{34.08}{1.0} & \stat{31.31}{0.4} & \stat{39.50}{1.8} & \statb{56.11}{5.1}\,\textcolor{gaingreen}{(+16.61)} \\

\rowcolor{green!5} ArtQuest & \stat{0.67}{0.1} & \stat{1.86}{0.1} & \stat{2.38}{0.3} & \stat{2.56}{0.2} & \stat{2.36}{0.1} & \stat{3.99}{0.3} & \stat{2.61}{0.2} & \stat{5.95}{0.2} & \statb{7.38}{2.3}\,\textcolor{gaingreen}{(+1.43)} \\

\rowcolor{green!5} AstroLLaVA & \stat{19.25}{0.6} & \stat{21.78}{0.7} & \stat{22.81}{0.7} & \stat{30.46}{1.0} & \stat{23.26}{0.8} & \stat{27.52}{0.8} & \stat{25.08}{0.6} & \stat{21.97}{1.7} & \statb{39.57}{1.5}\,\textcolor{gaingreen}{(+9.11)} \\

\rowcolor{green!5} GeoGrid-Bench & \stat{9.85}{0.3} & \stat{15.60}{0.6} & \stat{12.49}{0.4} & \stat{16.66}{0.6} & \stat{20.62}{0.7} & \stat{27.00}{0.2} & \stat{36.76}{0.5} & \stat{31.17}{0.6} & \statb{66.23}{4.7}\,\textcolor{gaingreen}{(+29.47)} \\
\bottomrule
\end{tabular}
}
\end{table*}

\section{Experiments}
\label{sec:exp}

We answer six RQs: \textbf{RQ1:} Does MMGraph-Agent outperform existing baselines across the three regimes? \textbf{RQ2:} Does each module of MMGraph-Agent work? \textbf{RQ3:} How cheap is the zero-API-cost multimodal hypergraph construction? \textbf{RQ4:} Does the cache-inspired memory bound the prompt under long-chain reasoning? \textbf{RQ5:} Does the offline-trained policy transfer to online retrieval without parameter updates? \textbf{RQ6:} How does MMGraph-Agent solve a hard representative multimodal case?

\subsection{Experimental Setup}

\noindent\textbf{Datasets and Evaluation Metrics.} We evaluate on \textbf{23 multimodal QA datasets} grouped by regime (\S\ref{app:dataset}): \textbf{(i) Knowledge-Only (K-only)}: DocBench~\citep{DocBench}, MMLongBench-Doc~\citep{MMLongBench-Doc}, ViDoSeek~\citep{ViDoSeek}, 2Wiki~\citep{2WikiMultiHopQA}, HotpotQA~\citep{HotpotQA}, Musique~\citep{Musique}, NQ~\citep{NQ}, PopQA~\citep{PopQA}, TriviaQA~\citep{TriviaQA}; \textbf{(ii) WebSearch-Only (Web-only)}: FVQA~\citep{FVQA}, InfoSeek~\citep{InfoSeek}, MMSearch~\citep{MMSearch}, SimpleVQA~\citep{SimpleVQA}, LiveVQA~\citep{LiveVQA}; \textbf{(iii) Knowledge+WebSearch (K+Web)}: MedXpertQA~\citep{MedXpertQA}, FinMME~\citep{FinMME}, We-Math2.0~\citep{We-Math2.0}, R-Bench~\citep{R-Bench}, MicroVQA~\citep{MicroVQA}, EEmo-Bench~\citep{EEmo-Bench}, ArtQuest~\citep{ArtQuest}, AstroLLaVA~\citep{AstroLLaVA}, GeoGrid-Bench~\citep{GeoGrid-Bench}; ViDoSeek, PopQA, TriviaQA and all Web-only datasets are held out as OOD to probe generalization. The only headline metric is token-level F$_{1}$ (\%) against the gold answer~\citep{Graph-R1,HyperGraphRAG}, which doubles as the per-rollout reward $R_{F_{1}}$; per-regime, per-dataset and efficiency breakdowns are in Appendix~\ref{app:eval}.

\noindent\textbf{Baselines and Ablations.} All methods share one corpus and prompt (\S\ref{app:baseline}): \textbf{(a)} generation-only NaiveGeneration (NaiveGen)~\citep{LLMsurvey}; \textbf{(b)} offline RAG-based StandardMMRAG (StdMMRAG)~\citep{VisRAG} and RAG-Anything (RAG-Any)~\citep{RAG-Anything}; \textbf{(c)} training-based SFT~\citep{SFT}, GRPO~\citep{GRPO}, and MMSearch-R1~\citep{MMSearch-R1} learning a web-search policy; plus \emph{w/o Hypergraph}, \emph{w/o Memory}, \emph{w/o Policy}, and \emph{w/o Retrieval} each isolating one module of \S\ref{sec:method}. NaiveGen, StdMMRAG and RAG-Any have GPT-4.1-mini~\citep{GPT4} variants.

\noindent\textbf{Implementation Details.} Backbone: Qwen2.5-VL-3B~\citep{Qwen2.5} with a frozen CLIP~\citep{CLIP} encoder $f$ ($D{=}768$). Memory: per-bank retrieval budget $C{=}5$; evidence truncated to $1{,}200$ words, condensed to an ${\approx}300$-word digest per round, context cap $5{,}000$ tokens. GRPO: $G{=}8$, $\epsilon{=}0.2$, $\kappa{=}10^{-3}$ over $75$ steps on $600$ examples ($233$ from six K-only and $367$ multimodal with $425$ images from eight in-domain datasets; none OOD), on $8\!\times\!$A100 (80GB); F$_{1}$ is mean$\pm$std over three sampled evaluation runs on the same $640$-item pool (temperature $1.0$); full hyperparameters in Appendix~\ref{app:impl}.

\clearpage

\subsection{Main Results (RQ1)}
\label{sec:rq1}

As shown in Table~\ref{tab:main_results}, MMGraph-Agent-3B improves over the second-best baseline per dataset by $+11.6$ F$_{1}$ on average across all $23$ datasets, with the largest gains on FinMME ($+44.93$), GeoGrid-Bench ($+29.47$), and ViDoSeek ($+23.06$); RAG-based and training-based baselines trail it on every regime, and the gap widens on long-document corpora and on the Knowledge+WebSearch regime where evidence must be combined across offline and online sources within one trajectory. On the eight held-out OOD datasets, it still leads the strongest published agent MMSearch-R1 by $+13.0$ F$_{1}$, indicating that the offline hypergraph and the shared retrieval schema generalize to distributions unseen during training, and it ranks first on all eight $\dagger$ (OOD) rows of Table~\ref{tab:main_results}.

\subsection{Ablation Study (RQ2)}
\label{sec:rq2}

\begin{figure*}[h]
\centering
\footnotesize
\begin{subfigure}[b]{0.33\linewidth}
  \centering
  \includegraphics[height=3.5cm]{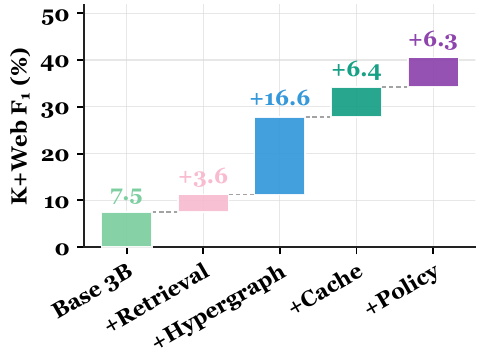}
  \vspace{-4mm}
  \caption{Components}
  \label{fig:rq2b}
\end{subfigure}\hfill
\begin{subfigure}[b]{0.33\linewidth}
  \centering
  \includegraphics[height=3.5cm]{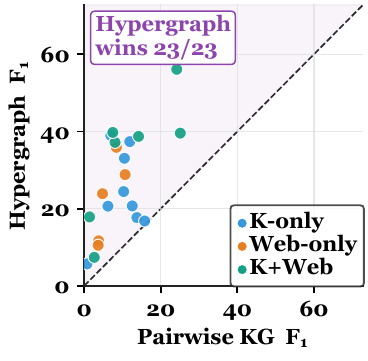}
  \vspace{-1mm}
  \caption{Representations}
  \label{fig:rq2c}
\end{subfigure}\hfill
\begin{subfigure}[b]{0.33\linewidth}
  \centering
  \includegraphics[height=3.5cm]{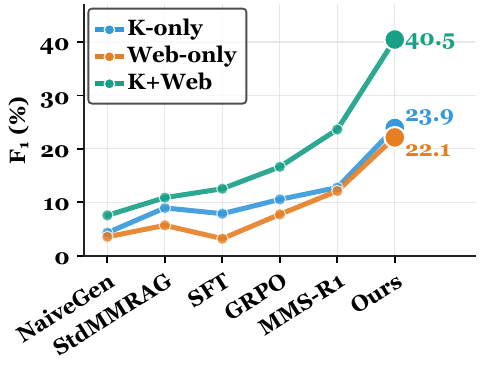}
  \vspace{-4mm}
  \caption{Regimes}
  \label{fig:rq2d}
\end{subfigure}
\vspace{-1mm}
\caption{(a) cumulative component waterfall on K+Web; (b) multimodal hypergraph against a pairwise KG, one point per dataset; (c) per-regime F$_{1}$ for the main baseline families and for Ours.}
\label{fig:rq2}
\end{figure*}

Figure~\ref{fig:rq2} ablates MMGraph-Agent along three complementary axes: per-module contribution (\textbf{a}), multimodal hypergraph vs.\ pairwise KG (\textbf{b}), and per-regime gain decomposition across baselines (\textbf{c}).

The waterfall in Figure~\ref{fig:rq2b} decomposes the K+Web F$_{1}$ from Base $7.5$ to Full $40.5$ into $+3.6$ retrieval, $+16.6$ hypergraph, $+6.4$ cache memory, and $+6.3$ policy learning (gains rounded), so the hypergraph contributes the most (roughly half of the ${\approx}33$-point gain) and every \S\ref{sec:method} module helps. Against a pairwise-edge knowledge graph built from the same entities and corpus, the multimodal hypergraph wins on $\mathbf{23/23}$ datasets (Figure~\ref{fig:rq2c}, one point per dataset), suggesting that $n$-ary hyperedges are the more effective primitive for cross-modal reasoning. The regime breakdown in Figure~\ref{fig:rq2d} shows MMGraph-Agent reaching $23.9/22.1/40.5$ on K-only / Web-only / K+Web, beating MMSearch-R1 by $+11.2 / +10.0 / +16.9$ F$_{1}$, with the largest margin on K+Web, where offline and online evidence must be combined, consistent with the per-dataset gains on the K+Web rows of Table~\ref{tab:main_results}.

\subsection{Analysis of Construction Cost (RQ3)}
\label{sec:rq3}

\begin{wrapfigure}[12]{r}{0.42\linewidth}
\centering
\vspace{-4mm}
\includegraphics[width=\linewidth]{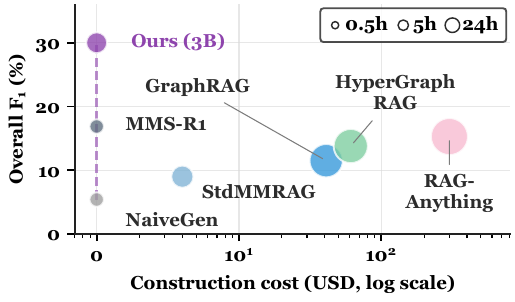}
\caption{Cost--performance Pareto on the same corpus; bubble area $\propto$ build time.}
\label{fig:rq3}
\end{wrapfigure}
As shown in Figure~\ref{fig:rq3}, we compare the construction cost and downstream F$_{1}$ of several multimodal knowledge organizers on the same $92$K-chunk, $74$K-image corpus.

\textbf{Construction cost.}
MMGraph-Agent reaches an overall F$_{1}$ of $30.0$ at zero API cost and a build of only a few hours on commodity CPUs, whereas GraphRAG and RAG-Anything need paid LLM extraction and far longer builds. The reduction is enabled by replacing LLM-based extraction with a deterministic spaCy NER and a frozen CLIP encoder, which eliminates the API cost of LLM-based extraction without degrading downstream accuracy, as the Pareto frontier in Figure~\ref{fig:rq3} shows.

\textbf{Hypergraph scale and modality routing.}
The constructed hypergraph contains roughly $2$M nodes and $0.92$M hyperedges over $18$ of the $23$ datasets, while the $5$ Web-only datasets are evaluated without an offline hypergraph by design. At inference, the policy issues image search for $35$--$39\%$ of actions on K+Web data and $0$--$3\%$ on K-only data, indicating that the policy relies on text queries over the offline hypergraph when queries are text-centric, with visual content indexed as image units (\S\ref{sec:construction}).

\clearpage

\subsection{Cache-Inspired Memory on Long-Chain Episodes (RQ4)}
\label{sec:rq4}

\begin{figure}[h]
\centering
\footnotesize
\vspace{-1mm}
\begin{subfigure}[t]{0.32\linewidth}
  \centering
  \includegraphics[width=\linewidth]{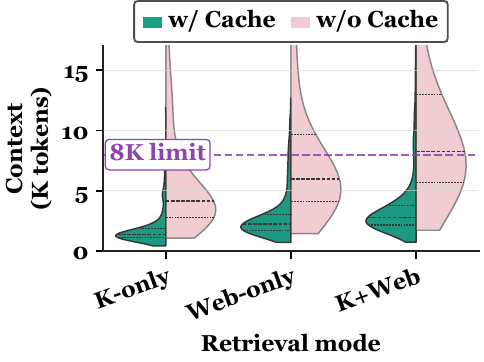}
  \vspace{-4mm}
  \caption{Context distribution}
  \label{fig:rq4a}
\end{subfigure}\hfill
\begin{subfigure}[t]{0.32\linewidth}
  \centering
  \includegraphics[width=\linewidth]{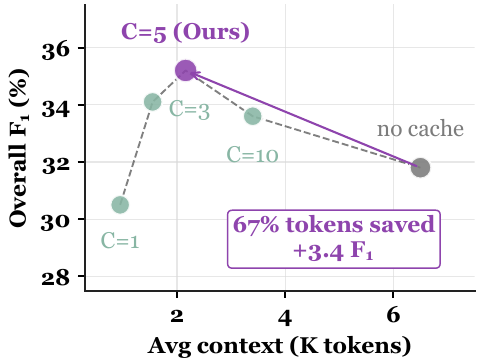}
  \vspace{-4mm}
  \caption{Cache-capacity Pareto}
  \label{fig:rq4b}
\end{subfigure}\hfill
\begin{subfigure}[t]{0.32\linewidth}
  \centering
  \includegraphics[width=\linewidth]{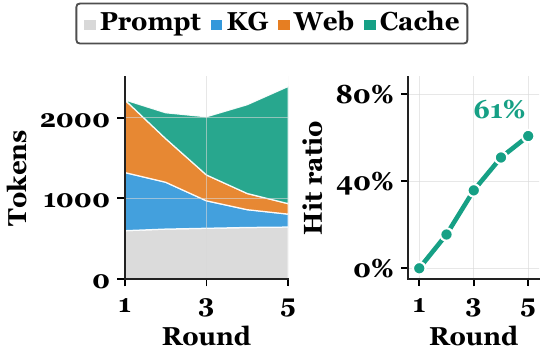}
  \vspace{-4mm}
  \caption{Per-round behavior}
  \label{fig:rq4c}
\end{subfigure}
\vspace{-1mm}
\caption{(a) context with/without L1 cache by regime; (b) cache-capacity Pareto across $C\in\{1,3,5,10,\infty\}$ ($\infty$: no cache); (c) per-round tokens and cache-hit ratio over rounds $1$--$5$.}
\label{fig:rq4}
\vspace{-1mm}
\end{figure}

As shown in Figure~\ref{fig:rq4}, we test whether the cache-inspired memory keeps the prompt bounded under long-chain reasoning of $3$ to $5$ rounds, where context overflow is the main failure mode. Long-chain trajectories are collected separately, since the 3B policy converges to one retrieval per query at the default budget (Figure~\ref{fig:rq5b}), too few rounds to stress the two-tier L1/L2 memory of \S\ref{sec:memory}.

\textbf{Bounded context and capacity sweep.}
Figure~\ref{fig:rq4a} shows that with the L1 cache per-query context stays mostly around $1$K--$3$K tokens across all three retrieval regimes, whereas without the L1 cache, a non-trivial fraction of trajectories, mostly on K+Web, exceeds the $8$K window. Sweeping the per-bank budget $C\in\{1,3,5,10,\infty\}$ in Figure~\ref{fig:rq4b}, the default $C{=}5$ lies on the Pareto frontier, saving $67\%$ tokens against the no-cache configuration while gaining $+3.4$ F$_{1}$, which motivates $C{=}5$ in \S\ref{sec:memory}.

\textbf{Per-round token economy.}
Figure~\ref{fig:rq4c} decomposes per-round context into prompt, KG-evidence, web-evidence and cache-hit components. The cache-hit share grows from $0\%$ at round~$1$ to roughly $61\%$ at round~$5$, so later rounds reuse evidence already cached by earlier rounds rather than issuing new tool calls, which bounds both the per-round prompt and the cumulative latency per episode.

\subsection{Analysis of Offline-to-Online Policy Transfer (RQ5)}
\label{sec:rq5_analysis}

\begin{figure}[h]
\centering
\footnotesize
\vspace{-1mm}
\begin{subfigure}[t]{0.34\linewidth}
  \centering
  \includegraphics[width=\linewidth]{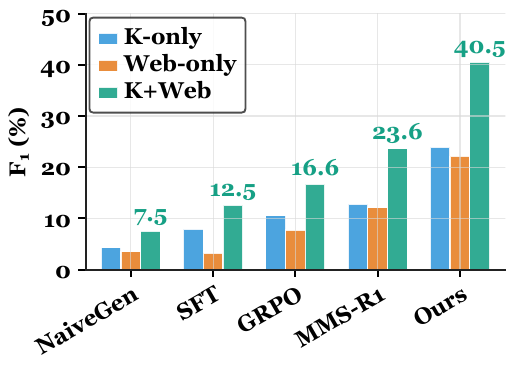}
  \vspace{-4mm}
  \caption{Per-regime unification gain}
  \label{fig:rq5a}
\end{subfigure}\hfill
\begin{subfigure}[t]{0.64\linewidth}
  \centering
  \includegraphics[width=\linewidth]{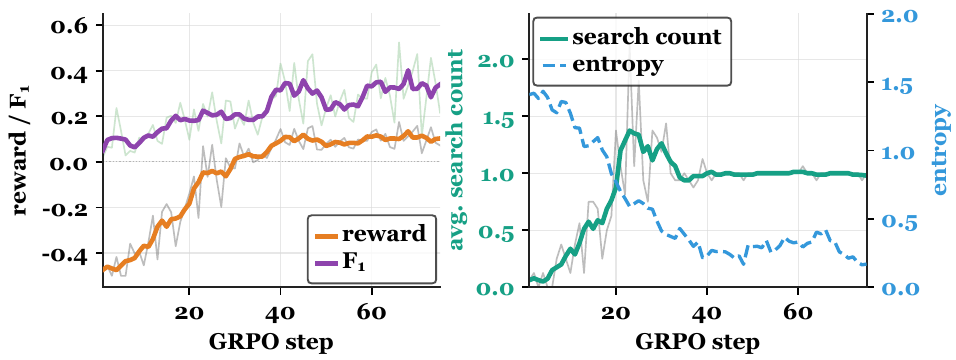}
  \vspace{-4mm}
  \caption{Training dynamics across $75$ GRPO steps}
  \label{fig:rq5b}
\end{subfigure}
\vspace{-1mm}
\caption{(a) per-regime F$_{1}$ for four 3B baselines and Ours; (b) 75-step GRPO trajectory of reward, F$_{1}$, search count, and entropy (thin lines: raw values; bold lines: 5-step moving averages).}
\label{fig:rq5}
\end{figure}

As shown in Figure~\ref{fig:rq5}, we decompose the trained 3B policy by retrieval regime and inspect its training dynamics across the full $75$ GRPO steps on $600$ in-domain examples, with web search disabled.

\textbf{Per-regime gains and action mix.}
Figure~\ref{fig:rq5a} reports per-regime F$_{1}$ for four 3B baselines and Ours, and the largest gain over MMSearch-R1 of $+16.9$ F$_{1}$ falls on K+Web. On K+Web the policy combines $\langle$\texttt{text\_search}$\rangle$ and $\langle$\texttt{search}$\langle$\texttt{img}$\rangle\rangle$ actions, while on K-only it issues text search for $97$ to $100\%$ of actions, and on Web-only it issues text search for $63$ to $68\%$ of actions.

\textbf{Training dynamics.}
Figure~\ref{fig:rq5b} shows reward improving from $-0.45$ at step~$1$ to roughly $+0.07$ at step~$75$, and per-batch F$_{1}$ rising from about $7$ to about $33$. The corresponding evaluation F$_{1}$ on the evaluation pool reaches $30.0$, the overall F$_{1}$ plotted in Figure~\ref{fig:rq3}. The search count grows from near zero to roughly $1.0$ by step~$25$ and stabilizes through step~$75$, while policy entropy decreases. The policy first learns when to emit a search tag, and then refines what to query in that call.

\clearpage

\subsection{Case Study (RQ6)}
\label{sec:rq6}

\begin{table*}[h]
  \centering
  \footnotesize
  \setlength{\tabcolsep}{6pt}
  \renewcommand{\arraystretch}{1.18}
  \caption{Case study on MedXpertQA sample 244. NaiveGen, MMSearch-R1 and MMGraph-Agent traces are all taken directly from the same evaluation logs on this sample. F$_{1}$ is reported per sample, with the dataset-level average shown in parentheses for reference.}
  \label{tab:casestudy}
  \begin{minipage}{\textwidth}
    \begin{minipage}[t]{0.82\textwidth}\vspace{0pt}\footnotesize
      \textbf{Question (MedXpertQA \#244).} A 71-year-old female presents with acute monocular vision loss in her right eye on awakening; no pain. PMH: myocardial infarction, hypertension, osteoporosis. BP 145/86\,mmHg, pupils symmetric and reactive. The fundus of the right eye is shown; the left fundus is normal. \emph{Choices:} (A) Retinal detachment\,/ (B) Central retinal artery occlusion\,/ (C) Retinal vein occlusion\,/ (D) Closed-angle glaucoma\,/ (E) Papilledema.\\[2pt]
      \textbf{Ground truth:} (C) Retinal vein occlusion.
    \end{minipage}\hfill
    \begin{minipage}[t]{0.12\textwidth}\vspace{0pt}\raggedleft
      \hspace*{-4mm}\fbox{\includegraphics[width=\linewidth]{figs/case_fundus_input.jpg}}\\
      \hspace*{-4mm}{\scriptsize\textit{Right-eye fundus.}}
    \end{minipage}
  \end{minipage}\\[6pt]
  \begin{tabular}{@{}>{\raggedright\arraybackslash}p{0.16\textwidth} >{\raggedright\arraybackslash}p{0.79\textwidth}@{}}
    \toprule
    \textbf{Method} & \textbf{Trace} \\
    \midrule

    \textbf{NaiveGen}\newline Qwen2.5-VL-3B\newline no retrieval\newline\textcolor{tagred}{F$_{1}$\,=\,0.50}\,\textcolor{gray}{\scriptsize(avg.\,0.028)}
    &
    \textcolor{tagblue}{\texttt{<reason>}}\,71-y/o female, acute monocular vision loss; normal pupillary reflexes consistent with vascular occlusion. \textbf{The fundus in Figure~A shows a characteristic appearance of CRAO: a narrowed retinal artery with `beading' and a pale optic disc.} \textcolor{tagred}{\textbf{[hallucinated: absent from the actual fundus]}}\,\textcolor{tagblue}{\texttt{</reason>}}\;\;\textcolor{tagred}{\texttt{<answer>}}\,\textbf{(B) Central retinal artery occlusion}\,\textcolor{tagred}{\texttt{</answer>}}\quad\textcolor{tagred}{\scriptsize$\boldsymbol{\times}$\,wrong; F$_{1}$\,=\,0.50 (shared tokens ``retinal'', ``occlusion'').}
    \\
    \midrule

    \textbf{MMSearch-R1}\newline Qwen2.5-VL-3B\newline RL-tuned web agent\newline (text + Lens, no cache)\newline\textcolor{tagred}{F$_{1}$\,=\,0.50}\,\textcolor{gray}{\scriptsize(avg.\,0.193)}
    &
    \textcolor{tagorange}{\texttt{<text\_search>}}\,\emph{CRAO vs RVO elderly painless monocular vision loss}\,\textcolor{tagorange}{\texttt{</text\_search>}}\;\;\textcolor{tagteal}{\texttt{<information>}}\,\emph{``CRAO classically presents with sudden, painless monocular vision loss in the elderly\,\dots''}\,\textcolor{tagteal}{\texttt{</information>}}\;\textcolor{gray}{\scriptsize[anchors symptom-driven prior on CRAO]}
    \par\smallskip
    \textcolor{tagorange}{\texttt{<search><img>}}\,\raisebox{-0.36\height}{\fbox{\includegraphics[height=8mm]{figs/case_fundus_input.jpg}}}\,\textcolor{tagorange}{\texttt{</search>}}\;\;\textcolor{tagteal}{\texttt{<information>}}\,\emph{``[Lens] The image likely depicts a retina with evidence of retinal vein occlusion (RVO)\,\dots''}\,\textcolor{tagteal}{\texttt{</information>}}\;\textcolor{gray}{\scriptsize[contradicts the text retrieval]}
    \par\smallskip
    \textcolor{tagblue}{\texttt{<reason>}}\,Two retrievals conflict; the symptom triad is the canonical prior, weigh text more.\,\textcolor{tagblue}{\texttt{</reason>}}\;\;\textcolor{tagred}{\texttt{<answer>}}\,\textbf{(B) Central retinal artery occlusion}\,\textcolor{tagred}{\texttt{</answer>}}\quad\textcolor{tagred}{\scriptsize$\boldsymbol{\times}$\,wrong; without cache-grounded reward to trust visual evidence, the policy anchors on the textbook prior.}
    \\
    \midrule

    \rowcolor{MMGraph!7}
    \textbf{MMGraph-Agent} (Ours)\newline Qwen2.5-VL-3B\newline hypergraph\,+\,L1/L2 cache\newline +\,graph/web policy\newline\textcolor{tagteal}{\textbf{F$_{1}$\,=\,1.00}}\,\textcolor{gray}{\scriptsize(avg.\,0.372)}
    &
    \textcolor{tagblue}{\texttt{<reason>}}\,\emph{Multiple vascular differentials (CRAO, RVO, detachment) match the symptoms; the fundus is the discriminating signal.}\,\textcolor{tagblue}{\texttt{</reason>}}\;\;\textcolor{tagorange}{\texttt{<search><img>}}\,\raisebox{-0.36\height}{\fbox{\includegraphics[height=8mm]{figs/case_fundus_input.jpg}}}\,\textcolor{tagorange}{\texttt{</search>}}
    \par\smallskip
    \fcolorbox{MMGraph!30}{MMGraph!4}{\parbox{0.96\linewidth}{\centering\scriptsize\textit{cache routing:}\,\colorbox{tagblue!18}{\strut~L2 warm-up~}\,$\to$\,\colorbox{MMGraph!22}{\strut~\textbf{multimodal hypergraph}~}\,$+$\,\colorbox{tagorange!22}{\strut~web search~}\,$\to$\,\colorbox{tagteal!22}{\strut~cache write~}}}
    \par\smallskip
    \textcolor{tagteal}{\texttt{<information>}}\,\emph{``[Search Results] The images in the provided figure likely depict a retina with evidence of retinal vein occlusion (RVO)\,\dots\,Since the 71-year-old female presented with acute monocular vision loss in her right eye, combined with waking sensations without pain, the most probable diagnosis is (C) Retinal vein occlusion.''}\,\textcolor{tagteal}{\texttt{</information>}}\;\textcolor{gray}{\scriptsize[digest]}
    \par\smallskip
    \textcolor{tagblue}{\texttt{<reason>}}\,\emph{Visual classification matches (C); CRAO prior never enters the chain.}\,\textcolor{tagblue}{\texttt{</reason>}}\;\;\textcolor{tagred}{\texttt{<answer>}}\,\textbf{Retinal vein occlusion}\,\textcolor{tagred}{\texttt{</answer>}}\quad\textcolor{teal!60!black}{\scriptsize$\checkmark$\,exact match; \texttt{search\_count}\,=\,1; one image-grounded round was both necessary and sufficient.}
    \\
    \bottomrule
  \end{tabular}
\end{table*}

Table~\ref{tab:casestudy} reports the trace. NaiveGen hallucinates fundus features and outputs (B) CRAO. MMSearch-R1 retrieves both the textbook CRAO passage and the correct Lens RVO classification but anchors on the textual evidence and gives the same wrong answer. MMGraph-Agent queries the image first; the cache routes through the multimodal hypergraph and the augmenting web channel, and the resulting evidence digest yields the correct diagnosis in one round, raising F$_{1}$ from 0.50 to 1.00.

\section{Conclusion}

In this work, we introduce MMGraph-Agent, a unified multimodal RAG framework that addresses three problems of current systems: costly knowledge organization, long-context retrieval, and the structural split between offline and online sources. It pairs a zero-API-cost multimodal hypergraph, a cache-inspired L1/L2 memory, and a GRPO-trained policy emitting retrieval and answer tokens over one shared action space. On $23$ multimodal QA datasets MMGraph-Agent outperforms generation-only, RAG-based, and training-based baselines, and ablations show that the three modules play complementary roles. Multilingual encoders and learned memory constants are left for future work.

\clearpage
\raggedbottom

\section*{Acknowledgments}
This work is supported by the National Research Foundation, Singapore, under its Frontier Competitive Research Programme (NRF-F-CRP-2024-0005). Any opinions, findings and conclusions or recommendations expressed in this material are those of the authors and do not reflect the views of National Research Foundation, Singapore. The authors declare no competing interests.

\begingroup\setlength{\bibsep}{9.1pt}
\bibliography{custom}{}
\bibliographystyle{plain}
\endgroup

\clearpage
\flushbottom
\appendix

\section*{Appendix}
\addcontentsline{toc}{section}{Appendix}

\section{MMGraph-Agent Operators and Prompts}\label{app:prompts}

This appendix lists every deterministic operator and LLM prompt of MMGraph-Agent, reproduced verbatim from the released code (\texttt{\{x\}} marks a runtime placeholder) in the prompt-card style of HyperGraphRAG~\citep{HyperGraphRAG} and grouped by the three modules of \S\ref{sec:method}. Modules~1 and~2 invoke no external LLM, so we show their core operators; only Module~3 prompts the unified GRPO-trained policy $\pi_{\theta}$.

\subsection{Zero-API-Cost Multimodal Hypergraph Construction (Module 1)}

\noindent{\textbf{Operator O1: Sentence-Level Multimodal Extractor} ($\pi_{\mathrm{ext}}$, \S\ref{sec:construction}).} spaCy splits each parsed chunk into sentences and turns every sentence into an $n$-ary hyperedge over its named entities; no LLM is invoked, so extraction incurs zero API cost and runs entirely on local CPU hardware.
\begin{shaded}
\scriptsize
\begin{verbatim}
# operate.py: zero-API-cost extractor running under ProcessPoolExecutor
def _mmextractor_worker(text: str):
    global _NLP
    doc = _NLP(text)                         # spaCy en_core_web_lg, frozen
    results = []
    for sent in doc.sents:
        ents = [ent.text for ent in sent.ents]
        if ents:
            results.append({
                "sentence": sent.text.strip(),
                "entities": list(dict.fromkeys(ents))
            })
    return results
\end{verbatim}
\end{shaded}

\noindent{\textbf{Operator O2: CLIP Embedding and FAISS Flat Index} ($f$, $\mathcal{G}_{H}^{m}$, \S\ref{sec:construction}).} Entities, hyperedges, images and queries share frozen CLIP ($D{=}768$) and flat FAISS inner-product indices: simple insertion, single-call exact Top-$k$ search over each bank, free of the recall loss of approximate indices.
\begin{shaded}
\scriptsize
\begin{verbatim}
# mmgraph.py: shared CLIP embedding bank for entity / hyperedge / image
embedding_model = SentenceTransformer('clip-ViT-L-14')        # d = 768
embeddings = model.encode(corpus_entity_content[corpus_source], batch_size=batch_size)
index = faiss.index_factory(dim, 'Flat', faiss.METRIC_INNER_PRODUCT)
index.add(corpus_numpy)                                       # append-only insertion
# search_utils.py: single exact Top-k call
_, ids = index.search(query_embedding, top_k)
\end{verbatim}
\end{shaded}

\subsection{Cache-Inspired Multimodal Hypergraph Memory (Module 2)}

\noindent{\textbf{Operator O3: Two-Tier Warm-Up Retrieval} (Eq.~(4) of \S\ref{sec:memory}).} Every round consults L2 (the hypergraph store, plus the web when enabled), admits new evidence into the per-trajectory L1 memory and serves the merged Top-$C$ from L1, so evidence retrieved in earlier rounds stays reusable.
\begin{shaded}
\scriptsize
\begin{verbatim}
# mem_server.py: L2 warm-up -> L1 serve on every round
kg_results = http_requests.post(f"http://localhost:{KG_PORT}/search",
    json={"queries": query_str}).json()       # L2: persistent hypergraph store
rag_memory.insert(unique_contexts)            # admit evidence into L1 memory
entity_match    = faiss_search(index_entity[traj_hash],
                               corpus_entity[traj_hash],    embeddings, 5)
hyperedge_match = faiss_search(index_hyperedge[traj_hash],
                               corpus_hyperedge[traj_hash], embeddings, 5)
result = rag_memory.query(query,              # merged Top-k served from L1
    param=QueryParam(only_need_context=True, top_k=5),
    entity_match=entity_match, hyperedge_match=hyperedge_match, ...)
\end{verbatim}
\end{shaded}

\noindent{\textbf{Operator O4: Digest-Bounded Prompt Admission} (Eq.~(5) of \S\ref{sec:memory}).} Evidence is truncated to $L_{\mathrm{max}}$ words and condensed into a digest of about $300$ words; the digest is admitted and rollout halts once the prompt reaches $L_{\mathrm{cap}}$ tokens, so the context stays bounded regardless of the number of rounds.
\begin{shaded}
\scriptsize
\begin{verbatim}
# rollout.py: bounded per-round admission into the prompt
retrieved_data = [item['<knowledge>']
                  for item in json.loads(res[index])['results'][:1]]
if len(corpus_text.split()) > 1200:           # truncate evidence payload
    corpus_text = ' '.join(corpus_text.split()[:1200]) + '...'
# ... the policy condenses <information> into an ~300-word digest ...
if len(vllm_inputs[i_gen]['prompt_token_ids']) >= 5000:
    idx_to_remove.append(i_gen)               # hard context cap: stop generating
\end{verbatim}
\end{shaded}

\subsection{Offline-Trained, Online-Deployed Policy Learning (Module 3)}

\noindent{\textbf{Prompt P0: Digest Request} (O4, \texttt{rollout.py}).} Sent after every retrieval; the reply is the digest $g_{t}$.
\begin{shaded}
\scriptsize
\begin{verbatim}
Searched results: <information>{retrieved evidence}</information>
Original user's question: {question}
Please provide an approximately 300-word summary of the current information related
to current query in English.
\end{verbatim}
\end{shaded}

\noindent{\textbf{Prompt P1: Initial Reasoning Prompt} (\S\ref{sec:rl}, Table~\ref{tab:mmgraph-agent-template}).} Issued at round $t{=}0$ (\texttt{dataset.py}); the policy reasons inside \texttt{<reason>} and then chooses image search, text search or the final answer, with the same prompt for every retrieval regime and identically at both training and inference time.
\begin{shaded}
\scriptsize
\begin{verbatim}
You are a helpful assistant. Answer the given question based on the provided images. You can
query from knowledge base provided to you to answer the question. You can query knowledge as
many times as you want. You must first conduct reasoning inside <reason>...</reason>. You
have two kinds of query tools, image search and text search. If you need to use image
search, you can append the string <search><img></search> behind reasoning. Format:
<reason>.....generate your reasoning here...</reason> <search><img></search>. This will
trigger a Google Lens search using the original image to retrieve relevant information that
can help you confirm the visual content. If you need to use text search, you can set a
concise and specific query, and output it in the format <text_search>your query
here</text_search> behind reasoning. Format: <reason>...generate your reasoning
here...</reason> <text_search>...generate your query here...</text_search>. The text search
tool will then use Google Search to return relevant information based on your query. When
you have the final answer, you can output the answer inside <answer>...</answer> behind
reasoning. Format: <reason>...generate your reasoning here...</reason> <answer>...generate
your answer here...</answer>. Please generate in English.
Here is the image and the question:
<image> {question}
\end{verbatim}
\end{shaded}

\noindent{\textbf{Prompt P2: Continuation with Image Input} (\S\ref{sec:rl}, \texttt{rollout.py}).} Appended to every retrieval result when the query carries an image: the evidence (or its digest) is wrapped in \texttt{<information>}, the question is restated, and all three actions remain available to the policy in the next round.
\begin{shaded}
\scriptsize
\begin{verbatim}
Searched results: <information>{evidence or digest}</information>
Original user's question: {question}
Based on information above, you need to determine whether one more Google image or text
search is needed to retrieve additional information, or whether to answer the
question directly. You must first conduct reasoning inside <reason>...</reason>. You have
three choices: image search, text search, or answer. If you need to use image search, you
can append the string <search><img></search> behind reasoning. Format: <reason>...generate
your reasoning here...</reason> <search><img></search>. If you need to use text search, you
can set a concise and specific query, and output it in the format <text_search>your query
here</text_search> behind reasoning. Format: <reason>...generate your reasoning
here...</reason> <text_search>...generate your query here...</text_search>. If you have the
final answer, you can output the answer inside <answer>...</answer> behind reasoning.
Format: <reason>...generate your reasoning here...</reason> <answer>...generate your answer
here...</answer>. Please generate in English.
\end{verbatim}
\end{shaded}

\noindent{\textbf{Prompt P3: Continuation for Text-Only Input} (\S\ref{sec:rl}, \texttt{rollout.py}).} Used instead of P2 when the query has no image, so the image-search branch is dropped, leaving text search and the answer.
\begin{shaded}
\scriptsize
\begin{verbatim}
Searched results: <information>{evidence or digest}</information>
Original user's question: {question}
Based on information above, you need to determine whether one more Google text search is
needed to retrieve additional information to answer the question or to answer the question
directly. You must first conduct reasoning inside <reason>...</reason>. You have two
choices: text search, or answer. If you need to use text search, you can set a concise and
specific query, and output it in the format <text_search>your query here</text_search>
behind reasoning. Format: <reason>...generate your reasoning here...</reason>
<text_search>...generate your query here...</text_search>. If you have the final answer, you
can output the answer inside <answer>...</answer> behind reasoning. Format:
<reason>...generate your reasoning here...</reason> <answer>...generate your answer
here...</answer>. Please generate in English.
\end{verbatim}
\end{shaded}

\subsection{Format-Reward Grammar (Module 3, training-time)}

\noindent{\textbf{Operator O5: Schema Compliance Regex} ($\mathbb{1}_{\mathrm{fmt}}$, Eq.~(6) of \S\ref{sec:rl}).} Each segment must match one anchored pattern with exactly one open/close pair per active tag; only fully compliant trajectories get $\mathbb{1}_{\mathrm{fmt}}{=}1$ and avoid the penalty $-\lambda$, which steers the policy toward well-formed, parseable trajectories.
\begin{shaded}
\scriptsize
\begin{verbatim}
# score.py: anchored regex grammar checked per response segment
direct_answer_format     = r'^<reason>.*</reason>.*<answer>.*</answer>$'
call_image_search_format = r'^<reason>.*</reason>.*<search><img></search>$'
call_text_search_format  = r'^<reason>.*</reason>.*<text_search>.*</text_search>$'

def is_valid_text_search(response, call_text_search_format) -> bool:
    if not re.match(call_text_search_format, response, re.DOTALL):
        return False
    if response.count('<reason>') != 1 or response.count('</reason>') != 1:
        return False
    if response.count('<text_search>') != 1 or response.count('</text_search>') != 1:
        return False
    ...                       # likewise rejects <answer> and <search><img></search> tags
    return True
\end{verbatim}
\end{shaded}

\noindent{\textbf{Operator O6: Hit-Aware Trajectory Reward} ($r(\tau)$, Eq.~(6) of \S\ref{sec:rl}).} Implements $r(\tau)\!=\!\mathbb{1}[N_c{\geq}1]\,\eta(N_c)R_{F_1}-\lambda(1-\mathbb{1}_{\mathrm{fmt}}\mathbb{1}[N_c{\geq}1])$ with $\eta(N){=}1{-}\tfrac{2}{N+2}$ and $\lambda{=}0.5$.
\begin{shaded}
\scriptsize
\begin{verbatim}
# score.py: per-trajectory reward, lambda = 0.5 (deepcopy omitted)
def compute_score(prediction, ground_truth, raw_response_str):
    answer = extract_solution(prediction=prediction[-1])
    score = 0
    if answer is not None:
        score = cal_f1([ground_truth], [answer])
    format_score, search_count = format_reward(prediction)
    if format_score == 1 and search_count >= 1:
        overall_score = score * (1 - 2 / (search_count + 2))
    elif format_score == 0 and search_count >= 1:
        overall_score = -0.5 + score * (1 - 2 / (search_count + 2))
    else:
        overall_score = -0.5
\end{verbatim}
\end{shaded}

\textbf{Reproducibility.} Operators O1--O6, prompts P1--P3, the rollout schema of Table~\ref{tab:mmgraph-agent-template} and the GRPO objective of Eq.~(7) fully specify the runtime. All generations sample with temperature $1.0$ and top-$p\,{=}\,1.0$ (vLLM) on Qwen2.5-VL-3B~\citep{Qwen2.5}, with at most $2{,}048$ new tokens per turn and $4{,}096$ per trajectory; O1--O2 run once over $\mathcal{K}$ (a few CPU hours for the $92$K-chunk suite at zero API cost) and O3--O4 run in the policy process on every rollout step, without any external LLM service.

\clearpage

\section{Proofs and Supporting Analyses}\label{app:proof}

This appendix supports Propositions~1--3 stated in \S\ref{sec:method}: Proposition~1 (zero-API-cost construction with comparable aggregate coverage, \S\ref{sec:construction}), Proposition~2 (bounded prompt with full reachability, \S\ref{sec:memory}), and Proposition~3 (offline-trained policy deployed online, \S\ref{sec:rl}). Proposition~2 receives a self-contained proof; the cost and interface halves of Propositions~1 and~3 are proved formally, while their coverage and transfer halves are empirical claims supported by \S\ref{sec:rq3} and \S\ref{sec:rq5_analysis}. Notation follows \S\ref{sec:method}: $\mathcal{G}_{H}^{m}\!=\!(\mathcal{V}^{m},\mathcal{E}_{H})$ is the multimodal hypergraph populated by the parser--extractor pair $(\psi,\pi_{\mathrm{ext}})$ under the frozen encoder $f\!:\!\mathcal{X}\!\to\!\mathbb{R}^{D}$, $\mathcal{M}_{1}^{t}$ and $\mathcal{M}_{2}$ are the L1/L2 memory tiers queried under the shared retrieval budget $C$, and $\pi_{\theta}$ is the GRPO-trained policy that emits trajectories $\tau\!=\!(a_{0},\ldots,a_{T})$ over the unified action space $\mathcal{A}\!=\!\mathcal{A}_{\mathrm{txt}}\!\cup\!\mathcal{A}_{\mathrm{img}}\!\cup\!\mathcal{A}_{\mathrm{ans}}$ shared by training and inference.

\subsection{Proof of Proposition 1}\label{app:proof1}
\noindent\textbf{Proposition 1.} \textit{$\mathcal{G}_{H}^{m}$ attains entity coverage comparable to an LLM extractor at zero API cost.}\vspace{-2mm}
\begin{proof}
Let $\pi_{\mathrm{ext}}^{\mathrm{LLM}}$ denote any LLM-based entity extractor returning canonical surface forms over the type alphabet $\mathcal{Y}\!=\!\{\textit{person},\textit{org},\textit{geo},\textit{event},\textit{category}\}$ used in HyperGraphRAG~\citep{HyperGraphRAG}, and let $\nu\!:\!\Sigma^{*}\!\to\!\Sigma^{*}_{\mathcal{Y}}$ be the deterministic UPPER-CASED canonicalizer shared with \S\ref{sec:construction}, where $\Sigma^{*}_{\mathcal{Y}}$ is the type-agnostic upper-cased canonical string space. For a corpus $\mathcal{K}$ and any extractor $\pi$, define the global coverage operator and the per-call monetary cost as
\begin{equation}
\mathrm{Cov}(\pi)\;=\!\!\bigcup_{d\in\mathcal{K}}\!\bigcup_{u\in\psi(d)}\!\!\nu\!\bigl(\pi(u)\bigr)\;\subseteq\;\Sigma^{*}_{\mathcal{Y}},\qquad \Gamma(\pi,\mathcal{K})\;=\;\sum_{d\in\mathcal{K}}\sum_{u\in\psi(d)}\!c(\pi,u)\;\in\;\mathbb{R}_{\geq 0},
\label{eq:p1-defs}
\end{equation}
where $c(\pi,u)\!\in\!\mathbb{R}_{\geq 0}$ is the dollar charge billed by $\pi$ on the call $u$. The local extractor of \S\ref{sec:construction} factorizes through the frozen NER head and frozen CLIP embedder,
\begin{equation}
\pi_{\mathrm{ext}}(u)\;=\;\mathrm{NER}_{\mathrm{spaCy}}(u),\qquad \mathbf{E}_{\mathcal{V}}\!\bigl(\pi_{\mathrm{ext}}(u)\bigr)\;=\;\bigl\{f(v)\!:\!v\!\in\!\pi_{\mathrm{ext}}(u)\bigr\}\;\subset\;\mathbb{R}^{D},
\label{eq:p1-ext}
\end{equation}
both of which run on local hardware with parameters frozen at load time and issue no network call, so $c(\pi_{\mathrm{ext}},u)\!=\!0$ for every $u$, and the corpus-level cost telescopes to
\begin{equation}
\Gamma(\pi_{\mathrm{ext}},\mathcal{K})\;=\;\sum_{d\in\mathcal{K}}\!\sum_{u\in\psi(d)}\!\!0\;=\;0,\qquad \Gamma(\pi_{\mathrm{ext}}^{\mathrm{LLM}},\mathcal{K})\;=\;\Theta\!\bigl(|\mathcal{K}|\!\cdot\!\bar{J}\!\cdot\!c_{\mathrm{LLM}}\bigr)\;>\;0,
\label{eq:p1-cost}
\end{equation}
where $\bar{J}\!=\!\mathbb{E}_{d\sim\mathcal{K}}[J(d)]$ and $c_{\mathrm{LLM}}\!>\!0$ is the strictly positive per-call API price, so the LLM baseline pays a dollar charge that scales linearly with corpus size while $\pi_{\mathrm{ext}}$ pays nothing. The OntoNotes classes of spaCy \texttt{en\_core\_web\_lg} cover the named-entity types of $\mathcal{Y}$, and both extractors feed Eq.~\eqref{eq:p1-defs} through the same type-agnostic canonicalizer $\nu$, so the only remaining discrepancy between the two coverages is the per-unit recall gap
\begin{equation}
\Delta(\mathcal{K})\;=\;\frac{\bigl|\mathrm{Cov}(\pi_{\mathrm{ext}}^{\mathrm{LLM}})\!\setminus\!\mathrm{Cov}(\pi_{\mathrm{ext}})\bigr|}{\bigl|\mathrm{Cov}(\pi_{\mathrm{ext}}^{\mathrm{LLM}})\bigr|}\;\in\;[0,1],\qquad \mathrm{Cov}(\pi_{\mathrm{ext}})\;\xrightarrow{\Delta\to 0}\;\mathrm{Cov}(\pi_{\mathrm{ext}}^{\mathrm{LLM}}),
\label{eq:p1-gap}
\end{equation}
The cost side of Proposition~1 is exact: Eq.~\eqref{eq:p1-cost} holds by construction. The coverage side is an empirical claim rather than a formal result: we do not measure $\Delta(\mathcal{K})$ directly, and general-domain NER may differ from an LLM extractor on specialized vocabulary; \S\ref{sec:rq3} supports comparable aggregate coverage indirectly through the resulting graph scale and the downstream F$_{1}$ of Table~\ref{tab:main_results}, which is the precise sense in which ``comparable'' is used in the statement of Proposition~1.
\end{proof}

\subsection{Proof of Proposition 2}\label{app:proof2}
\noindent\textbf{Proposition 2.} \textit{The two-tier memory bounds the prompt uniformly over $T$ with $\mathcal{G}_{H}^{m}$ fully reachable.}\vspace{-2mm}
\begin{proof}
The L1 memory and the prompt evolve under Eqs.~(4)--(5): each round admits the fresh L2 (and web) evidence into L1 and appends one bounded digest to the conversation,
\begin{equation}
\mathcal{M}_{1}^{t}\;=\;\mathcal{M}_{1}^{t-1}\!\cup\!\mathcal{R}_{2}(q_{t})\!\cup\!\mathcal{R}_{\mathrm{on}}(q_{t}),\qquad \mathcal{M}_{1}^{0}\;=\;\varnothing,\qquad \mathrm{prompt}(s_{t+1})\;=\;\mathrm{prompt}(s_{t})\oplus g_{t},
\label{eq:p2-evolve}
\end{equation}
where the digest $g_{t}$ condenses the merged evidence block $\mathcal{K}^{*}_{q_{t}}$ (truncated to $L_{\mathrm{max}}$ words) and is itself bounded,
\begin{equation}
|g_{t}|\;\leq\;L_{\mathrm{dig}},\qquad \forall\,t\!\in\!\{0,\ldots,T-1\}.
\label{eq:p2-card}
\end{equation}
Writing $s_{t}\!=\!(q,v_{q},\mathcal{M}_{1}^{t-1},h_{t})$ for the agent state, an immediate induction on $t$, combined with the context threshold $L_{\mathrm{cap}}$ that halts the agent rollout once reached, yields the prompt bound
\begin{equation}
|\mathrm{prompt}(s_{t})|\;\leq\;\min\bigl(|\mathrm{prompt}(s_{0})|\!+\!t\!\cdot\!L_{\mathrm{dig}},\;L_{\mathrm{cap}}\bigr)\;\leq\;L_{\mathrm{cap}},\qquad \forall\,t\!\in\!\{0,\ldots,T\},
\label{eq:p2-bound}
\end{equation}
independent of the trajectory horizon $T$ and dependent on the admission constants $(L_{\mathrm{dig}},L_{\mathrm{cap}})$ alone. For reachability, the L2 backend stores the entire constructed hypergraph store and is queried by exact inner-product Top-$C$ under the flat FAISS index on every round,
\begin{equation}
\mathcal{M}_{2}\;=\;\mathcal{V}^{m}\!\cup\!\mathcal{E}_{H}\!\cup\!\mathcal{I},\qquad \mathcal{R}_{2}(q_{t})\;=\!\!\!\operatornamewithlimits{Top\text{-}C}_{x\in\mathcal{M}_{2}}\!\!\!\langle f(q_{t}),f(x)\rangle,
\label{eq:p2-l2}
\end{equation}
unconditionally, so that no gating step can mask any part of $\mathcal{M}_{2}$ from a retrieval action,
\begin{equation}
\mathcal{R}_{2}(q_{t})\;\subseteq\;\mathcal{M}_{1}^{t},\qquad \mathcal{K}^{*}_{q_{t}}\;=\!\!\!\operatornamewithlimits{Top\text{-}C}_{x\in\mathcal{M}_{1}^{t}}\!\!\!\langle f(q_{t}),f(x)\rangle.
\label{eq:p2-miss}
\end{equation}
Fix any node $x^{*}\!\in\!\mathcal{G}_{H}^{m}$ and choose $q_{t}\!\in\!\mathcal{X}$ whose CLIP encoding aligns with $f(x^{*})$ (the surface form for $x^{*}\!\in\!\mathcal{V}^{m}$, or the host sentence for $x^{*}\!\in\!\mathcal{E}_{H}$): the inner product $\langle f(q_{t}),f(x^{*})\rangle$ then ranks $x^{*}$ within the exact Top-$C$ of Eq.~\eqref{eq:p2-l2}, so $x^{*}\!\in\!\mathcal{R}_{2}(q_{t})\!\subseteq\!\mathcal{M}_{1}^{t}$, and a quantifier sweep over $q_{t}\!\in\!\mathcal{X}$ shows
\begin{equation}
\bigcup_{q_{t}\in\mathcal{X}}\!\mathcal{R}_{2}(q_{t})\;=\;\mathcal{V}^{m}\!\cup\!\mathcal{E}_{H}\!\cup\!\mathcal{I}\;\supseteq\;\mathcal{V}^{m}\!\cup\!\mathcal{E}_{H},
\label{eq:p2-reach}
\end{equation}
so every stored item remains recoverable on demand through L2 with no loss of accessibility, while the prompt never ingests more than one bounded digest per round. Combining Eq.~\eqref{eq:p2-bound} with Eq.~\eqref{eq:p2-reach} establishes Proposition~2, with both guarantees holding for every trajectory horizon $T$.
\end{proof}

\subsection{Proof of Proposition 3}\label{app:proof3}
\noindent\textbf{Proposition 3.} \textit{The source-agnostic interface deploys the offline-trained $\pi_{\theta}$ online without updates.}\vspace{-2mm}
\begin{proof}
Denote the training and inference distributions by $\mathcal{D}_{\mathrm{off}}$ and $\mathcal{D}_{\mathrm{on}}$ respectively, sharing the query marginal of $\mathcal{D}$ but differing in the cached evidence source,
\begin{equation}
\mathcal{K}^{*}_{q_{t}}\!\subseteq\!\mathcal{V}^{m}\!\cup\!\mathcal{E}_{H}\!\cup\!\mathcal{I}\;\;\text{under }\mathcal{D}_{\mathrm{off}}\;(\mathcal{R}_{\mathrm{on}}\!=\!\varnothing),\qquad \mathcal{K}^{*}_{q_{t}}\!\subseteq\!\mathcal{V}^{m}\!\cup\!\mathcal{E}_{H}\!\cup\!\mathcal{I}\!\cup\!\mathcal{S}(\mathcal{W})\;\;\text{under }\mathcal{D}_{\mathrm{on}}\;(\mathcal{R}_{\mathrm{on}}\!=\!\mathcal{S}(\mathcal{W})).
\label{eq:p3-source}
\end{equation}
Every evidence element enters the prompt only through the per-round digests of Eq.~\eqref{eq:p2-evolve}, and the templating functions depend on the retrieved evidence only through the digest tokens, giving the source-agnostic prompt
\begin{equation}
\mathrm{prompt}(s_{t})\;=\;T_{\mathrm{seed}}(q,v_{q},h_{t})\;\oplus\;\bigoplus_{i<t}T_{\mathrm{slot}}\!\bigl(g_{i}\bigr),
\label{eq:p3-prompt}
\end{equation}
where neither $T_{\mathrm{seed}}$ nor $T_{\mathrm{slot}}$ branches on the provenance of the evidence underlying $g_{i}$. Because the same encoder $f$ embeds entities, hyperedges and SerpAPI evidence, the inference embedding set lies in the training image $f(\mathcal{X})$, and the action set $\mathcal{A}$ is identical at both stages, so for every $s_{t}\!\in\!\mathrm{supp}(\mathcal{D}_{\mathrm{on}})$ there exists $s_{t}'\!\in\!\mathrm{supp}(\mathcal{D}_{\mathrm{off}})$ with $\mathrm{prompt}(s_{t})\!=\!\mathrm{prompt}(s_{t}')$, yielding the policy invariance
\begin{equation}
\pi_{\theta}\!\bigl(a\!\mid\!s_{t}\bigr)\;=\;\pi_{\theta}\!\bigl(a\!\mid\!s_{t}'\bigr),\qquad \forall\,a\!\in\!\mathcal{A},\;\forall\,s_{t}\!\in\!\mathrm{supp}(\mathcal{D}_{\mathrm{on}}).
\label{eq:p3-pi-inv}
\end{equation}
Letting $\theta^{*}\!=\!\arg\max_{\theta}\mathcal{J}_{\mathrm{GRPO}}(\theta;\mathcal{D}_{\mathrm{off}})$ be the GRPO-optimal parameters trained under $\mathcal{D}_{\mathrm{off}}$ via Eq.~(7), Eqs.~\eqref{eq:p3-prompt}--\eqref{eq:p3-pi-inv} show that enabling $\mathcal{R}_{\mathrm{on}}$ changes neither the action space, the prompt template, nor the embedding interface seen by $\pi_{\theta^{*}}$: the web extension is invisible to the policy's input--output contract. This is a structural statement about the evidence interface, not a distributional one: $\mathcal{D}_{\mathrm{on}}$ may shift the evidence distribution, no gradient ever flows from the online channel, and we make no invariance claim about $\mathcal{J}_{\mathrm{GRPO}}(\theta;\mathcal{D}_{\mathrm{on}})$. The transfer itself is therefore an empirical finding: \S\ref{sec:rq5_analysis} shows the same checkpoint scoring all three retrieval regimes unchanged, which is the precise content of Proposition~3, obtained with no online gradient step and no SerpAPI call during training.
\end{proof}

\textbf{Summary.} The three results separate what holds by construction from what is measured. The zero-cost side of Proposition~1 (Eq.~\eqref{eq:p1-cost}), the prompt bound and full reachability of Proposition~2 (Eqs.~\eqref{eq:p2-bound} and~\eqref{eq:p2-reach}), and the interface invariance of Proposition~3 (Eq.~\eqref{eq:p3-pi-inv}) are exact; the coverage side of Proposition~1 and the online transfer of Proposition~3 are empirical, supported by the graph scale and downstream F$_{1}$ of \S\ref{sec:rq3} and the unchanged checkpoint of \S\ref{sec:rq5_analysis}, respectively.

\clearpage

\section{MMGraph-Agent Algorithm Details}\label{app:algo}

This appendix expands \S\ref{sec:method} into executable pseudocode for MMGraph-Agent's three contributions, following the algorithm-card style of HyperGraphRAG~\citep{HyperGraphRAG}. Algorithm~\ref{alg:mmga-c1} populates the multimodal hypergraph $\mathcal{G}_{H}^{m}\!=\!(\mathcal{V}^{m},\mathcal{E}_{H})$ of \S\ref{sec:construction} from the static corpus $\mathcal{K}$ through the local parser--extractor pair $(\psi,\pi_{\mathrm{ext}})$ under the frozen CLIP encoder $f\!:\!\mathcal{X}\!\to\!\mathbb{R}^{D}$; Algorithm~\ref{alg:mmga-c2} realizes the cache-inspired two-tier memory of \S\ref{sec:memory} as a per-round controller that exposes the same Top-$C$ primitive on both tiers $\mathcal{M}_{1}^{t}$ and $\mathcal{M}_{2}$ and keeps the prompt bounded via digest admission; and Algorithm~\ref{alg:mmga-c3} fits the unified policy $\pi_{\theta}$ via the GRPO clipped surrogate of \S\ref{sec:rl} on offline-only rollouts, leaving the inference-time activation of $\mathcal{R}_{\mathrm{on}}\!=\!\mathcal{S}(\mathcal{W})$ as a free switch requiring no retraining or prompt change.

\textbf{Module 1: Zero-API-Cost Multimodal Hypergraph Construction.} Following \S\ref{sec:construction}, every document $d\!\in\!\mathcal{K}$ is first sliced by the rule-based parser $\psi$ into modality-tagged units $\{u_{j}^{\mu_{j}}\}$, and each textual unit is fed to the local extractor $\pi_{\mathrm{ext}}\!=\!\mathrm{NER}_{\mathrm{spaCy}}$ that emits the multiset of entities and the host sentence as the corresponding $n$-ary hyperedge, while image units are embedded directly by $f$ into the shared retrieval space. The frozen multimodal encoder $f$ embeds entities, hyperedges, images and queries into a single $D{=}768$ space, and a streaming flat FAISS inner-product index serves exact cross-modal Top-$k$ with fast append-only insertion. No paid LLM is invoked at any stage, so the entire entity layer of $\phi$ is built at \emph{zero API cost} and can be re-run locally on every newly arriving batch, extending the searchable index on ordinary commodity hardware at no API cost. Each hyperedge keeps its full host sentence as payload, so higher-order context stays within one retrievable unit.

\begin{algorithm}[H]
\small
\caption{Module~1: Zero-API-Cost Multimodal Hypergraph Construction}
\label{alg:mmga-c1}
\begin{algorithmic}[1]
\REQUIRE Static corpus $\mathcal{K}\!=\!\{d_{i}\}_{i=1}^{N}$, modality alphabet $\mathbb{M}$, frozen CLIP encoder $f$
\ENSURE Multimodal hypergraph $\mathcal{G}_{H}^{m}\!=\!(\mathcal{V}^{m},\mathcal{E}_{H})$ with embedding banks $(\mathbf{E}_{\mathcal{V}},\mathbf{E}_{\mathcal{E}_{H}},\mathbf{E}_{\mathcal{I}})$
\STATE Initialize $\mathcal{V}^{m}\!\leftarrow\!\varnothing,\ \mathcal{E}_{H}\!\leftarrow\!\varnothing,\ \mathbf{E}_{\mathcal{V}}\!\leftarrow\!\varnothing,\ \mathbf{E}_{\mathcal{E}_{H}}\!\leftarrow\!\varnothing,\ \mathbf{E}_{\mathcal{I}}\!\leftarrow\!\varnothing$
\FOR{each document $d\!\in\!\mathcal{K}$ in parallel}
    \STATE $\{u_{j}^{\mu_{j}}\}_{j=1}^{J(d)}\!\leftarrow\!\psi(d)$ \hfill \COMMENT{rule-based parser tags each unit by $\mu_{j}\!\in\!\mathbb{M}$}
    \FOR{each textual unit $u\!\in\!\psi_{\mathrm{txt}}(d)$}
        \STATE $V_{u}\!\leftarrow\!\pi_{\mathrm{ext}}(u)\!=\!\mathrm{NER}_{\mathrm{spaCy}}(u)$ \hfill \COMMENT{local entities, no LLM call}
        \STATE $\mathcal{V}^{m}\!\leftarrow\!\mathcal{V}^{m}\!\cup\!V_{u}$
        \IF{$|V_{u}|\!\geq\!1$}
            \STATE $e_{H}\!\leftarrow\!\mathrm{sentence}(u)$ as $n$-ary hyperedge over $V_{u}$ with the sentence as payload
            \STATE $\mathcal{E}_{H}\!\leftarrow\!\mathcal{E}_{H}\!\cup\!\{e_{H}\}$
        \ENDIF
    \ENDFOR
    \STATE $\mathbf{E}_{\mathcal{I}}\!\leftarrow\!\mathbf{E}_{\mathcal{I}}\!\cup\!\{f(u):u\!\in\!\psi(d),\,\mu(u)\!=\!\mathrm{image}\}$ \hfill \COMMENT{image units embedded directly}
\ENDFOR
\STATE $\mathbf{E}_{\mathcal{V}}\!\leftarrow\!\{f(v)\!:\!v\!\in\!\mathcal{V}^{m}\}$,\ \ $\mathbf{E}_{\mathcal{E}_{H}}\!\leftarrow\!\{f(e_{H})\!:\!e_{H}\!\in\!\mathcal{E}_{H}\}$ \hfill \COMMENT{frozen CLIP, $D{=}768$}
\STATE Insert $\mathbf{E}_{\mathcal{V}}\!\cup\!\mathbf{E}_{\mathcal{E}_{H}}\!\cup\!\mathbf{E}_{\mathcal{I}}$ into flat FAISS inner-product indices keyed on $\mathrm{sim}(x,x')\!=\!\langle f(x),f(x')\rangle$
\RETURN $\mathcal{G}_{H}^{m}\!=\!(\mathcal{V}^{m},\mathcal{E}_{H})$ with $(\mathbf{E}_{\mathcal{V}},\mathbf{E}_{\mathcal{E}_{H}},\mathbf{E}_{\mathcal{I}})$
\end{algorithmic}
\end{algorithm}

\textit{Complexity.} With $|\mathcal{K}|\!=\!N$ documents, $\bar{J}\!=\!\mathbb{E}_{d}[J(d)]$ units per document, and $\bar{n}$ entities per unit, the entity scan costs $\mathcal{O}(N\!\cdot\!\bar{J}\!\cdot\!\bar{n})$ CPU operations on spaCy that parallelize trivially across $\mathcal{K}$. Embedding $\mathcal{V}^{m}\!\cup\!\mathcal{E}_{H}$ is one frozen-CLIP forward pass per element, and flat FAISS insertion is lightweight per node, so the wall-clock for the $92$K-chunk evaluation corpus stays within a few CPU hours at zero API cost (\S\ref{sec:rq3}), strictly dominating the API-bound profile of LLM-based construction baselines.

\textbf{Module 2: Cache-Inspired Multimodal Hypergraph Memory.} Following \S\ref{sec:memory}, the offline retrieval channel $\mathcal{R}_{\mathrm{off}}$ is split across two tiers: the L2 backend $\mathcal{M}_{2}$ persistently retains every entity, hyperedge, and visual item, the L1 per-trajectory cache $\mathcal{M}_{1}^{t}$ accumulates the evidence retrieved so far, and both are queried by the same Top-$C$ inner-product primitive on $\mathcal{X}$. The policy $\pi_{\theta}$ itself emits the sub-query $q_{t}\!\in\!\mathcal{X}$ inside its retrieval action; the controller consults L2 (and, when enabled, the web channel) on every round, admits the fresh evidence into L1, and serves the merged evidence $\mathcal{K}^{*}_{q_{t}}$ from the unified L1 index. The evidence block is truncated to $L_{\mathrm{max}}$ words and condensed into a bounded digest before entering the conversation until the halting threshold $L_{\mathrm{cap}}$, so the prompt length is bounded uniformly in the horizon $T$, while every stored item remains fully reachable through the L2 backend. Starting from an empty L1 cache, the working set is warmed from the hypergraph rather than from policy priors, and later rounds re-serve the evidence already held in L1, with one budget for both tiers.

\begin{algorithm}[H]
\small
\caption{Module~2: Cache-Inspired Two-Tier Memory at Round $t$}
\label{alg:mmga-c2}
\begin{algorithmic}[1]
\REQUIRE State $s_{t}\!=\!(q,v_{q},\mathcal{M}_{1}^{t-1},h_{t})$, backend $\mathcal{M}_{2}\!=\!\mathcal{V}^{m}\!\cup\!\mathcal{E}_{H}\!\cup\!\mathcal{I}$, encoder $f$, budget $C$, bounds $(L_{\mathrm{max}},L_{\mathrm{dig}},L_{\mathrm{cap}})$
\ENSURE Round-$t$ cache $\mathcal{M}_{1}^{t}$ and digest $g_{t}$ appended to the prompt
\STATE $q_{t}\!\leftarrow\!\pi_{\theta}$'s retrieval action \hfill \COMMENT{sub-query emitted autoregressively, $q_{t}\!\in\!\mathcal{X}$}
\STATE $\mathcal{R}_{2}(q_{t})\!\leftarrow\!\mathrm{Top}\text{-}C_{x\in\mathcal{M}_{2}}\langle f(q_{t}),f(x)\rangle$ \hfill \COMMENT{L2 warm-up via flat FAISS index, every round}
\STATE $\mathcal{R}_{\mathrm{on}}(q_{t})\!\leftarrow\!\mathcal{S}(\mathcal{W},q_{t})$ if web search enabled, else $\varnothing$
\STATE $\mathcal{M}_{1}^{t}\!\leftarrow\!\mathcal{M}_{1}^{t-1}\!\cup\!\mathcal{R}_{2}(q_{t})\!\cup\!\mathcal{R}_{\mathrm{on}}(q_{t})$ \hfill \COMMENT{admit fresh evidence into the L1 memory}
\STATE $\mathcal{K}^{*}_{q_{t}}\!\leftarrow\!\mathrm{Top}\text{-}C_{x\in\mathcal{M}_{1}^{t}}\langle f(q_{t}),f(x)\rangle$ \hfill \COMMENT{merged evidence served from the unified L1 index}
\STATE Truncate $\mathcal{K}^{*}_{q_{t}}$ to $L_{\mathrm{max}}$ words; $g_{t}\!\leftarrow\!$ policy digest of $\mathcal{K}^{*}_{q_{t}}$, $|g_{t}|\!\leq\!L_{\mathrm{dig}}$
\STATE $\mathrm{prompt}(s_{t+1})\!\leftarrow\!\mathrm{prompt}(s_{t})\oplus g_{t}$; halt the trajectory if $|\mathrm{prompt}|\!\geq\!L_{\mathrm{cap}}$
\RETURN $(\mathcal{M}_{1}^{t},g_{t})$
\end{algorithmic}
\end{algorithm}

\textit{Complexity.} Each round costs one exact L2 Top-$C$ over the flat FAISS index, one insertion of the fresh evidence into the L1 memory, and one Top-$C$ serve from the L1 index, giving a per-round controller cost that does not grow with the prompt across the horizon $T$. Only the bounded digests enter the language-model prompt until the halting threshold $L_{\mathrm{cap}}$, so the prompt length stays bounded across the entire trajectory while the full $\mathcal{G}_{H}^{m}$ remains reachable through L2, matching the bound proved in Proposition~2 for every trajectory, every corpus size and all three retrieval regimes alike.

\textbf{Module 3: Offline-Trained, Online-Deployed Policy Learning.} Following \S\ref{sec:rl}, the policy $\pi_{\theta}$ samples actions $a_{t}\!\sim\!\pi_{\theta}(\cdot|s_{t})$ from the unified action space $\mathcal{A}\!=\!\mathcal{A}_{\mathrm{txt}}\!\cup\!\mathcal{A}_{\mathrm{img}}\!\cup\!\mathcal{A}_{\mathrm{ans}}$ in one autoregressive emission, with both retrieval channels routed through the cache so the policy emits a single retrieval token agnostic to its source. Training samples $G$ rollouts per query under web augmentation disabled, computes the hit-aware reward $r(\tau)\!=\!\mathbb{1}[N_{c}\!\geq\!1]\eta(N_{c})R_{F_{1}}\!-\!\lambda(1\!-\!\mathbb{1}_{\mathrm{fmt}}\mathbb{1}[N_{c}\!\geq\!1])$ with the smooth call-count regulator $\eta(N)\!=\!1\!-\!\tfrac{2}{N+2}$, and updates $\theta$ via the GRPO clipped surrogate anchored to a frozen reference $\pi_{\mathrm{ref}}$. At inference $\mathcal{R}_{\mathrm{on}}\!=\!\mathcal{S}(\mathcal{W})$ is enabled and the unified evidence interface lets the trained policy serve online retrieval without any further parameter updates or retraining.

\begin{algorithm}[H]
\small
\caption{Module~3: Offline-Only GRPO Training of $\pi_{\theta}$}
\label{alg:mmga-c3}
\begin{algorithmic}[1]
\REQUIRE Training set $\mathcal{D}$, hypergraph $\mathcal{G}_{H}^{m}$, frozen reference $\pi_{\mathrm{ref}}$, group size $G$, clip width $\epsilon$, KL coefficient $\kappa$
\ENSURE GRPO-optimal parameters $\theta^{*}$
\STATE Initialize $\theta\!\leftarrow\!\theta_{0}$ from the Qwen2.5-VL-3B backbone
\FOR{step $=\!1,\ldots,75$}
    \STATE Sample query $(q,v_{q},y^{*})\!\sim\!\mathcal{D}$ with $\mathcal{R}_{\mathrm{on}}\!=\!\varnothing$ (offline only)
    \FOR{$i\!=\!1,\ldots,G$ in parallel}
        \STATE Roll out $\tau_{i}\!\sim\!\pi_{\theta_{\mathrm{old}}}(\cdot|q,v_{q})$ via Algorithm~\ref{alg:mmga-c2} per round
        \STATE Compute $r(\tau_{i})\!=\!\mathbb{1}[N_{c}^{(i)}\!\geq\!1]\eta(N_{c}^{(i)})R_{F_{1}}^{(i)}\!-\!\lambda\bigl(1\!-\!\mathbb{1}_{\mathrm{fmt}}^{(i)}\!\cdot\!\mathbb{1}[N_{c}^{(i)}\!\geq\!1]\bigr)$
    \ENDFOR
    \STATE $\mu_{G}\!\leftarrow\!\tfrac{1}{G}\!\sum_{i}\!r(\tau_{i}),\ \sigma_{G}\!\leftarrow\!\mathrm{std}_{i}\,r(\tau_{i})$,\ \ $\hat{A}_{i}\!\leftarrow\!(r(\tau_{i})\!-\!\mu_{G})/(\sigma_{G}\!+\!\delta)$
    \STATE $\rho_{t}\!\leftarrow\!\pi_{\theta}(a_{t}|s_{t})/\pi_{\theta_{\mathrm{old}}}(a_{t}|s_{t})$ for every token of every $\tau_{i}$
    \STATE $\theta\!\leftarrow\!\theta\!+\!\alpha_{\mathrm{lr}}\!\cdot\!\nabla_{\theta}\!\Bigl[\!\sum_{i,t}\!\tfrac{1}{G|\tau_{i}|}\min\!\bigl(\rho_{t}\hat{A}_{i},\,\mathrm{clip}(\rho_{t},1\!\pm\!\epsilon)\hat{A}_{i}\bigr)\!-\!\kappa\,\mathrm{KL}(\pi_{\theta}\!\parallel\!\pi_{\mathrm{ref}})\Bigr]$
\ENDFOR
\RETURN $\theta^{*}\!=\!\theta$ \hfill \COMMENT{at inference, activate $\mathcal{R}_{\mathrm{on}}\!=\!\mathcal{S}(\mathcal{W})$ via the same cache interface}
\end{algorithmic}
\end{algorithm}

\textit{Complexity.} Each step issues $G$ rollouts whose cost is dominated by Algorithm~\ref{alg:mmga-c2} times the autoregressive decode of $\pi_{\theta}$; the gradient pass is one forward-backward over the GRPO surrogate at $\mathcal{O}(G\!\cdot\!\bar{|\tau|}\!\cdot\!|\theta|)$ per step, executed on $8\!\times\!$A100 (80GB) for $75$ steps with $G\!=\!8,\epsilon\!=\!0.2,\kappa\!=\!10^{-3}$. Because training is offline-only and $\mathcal{S}$ is not invoked during rollouts, the SerpAPI bill scales with the inference budget alone; combined with the source-agnostic evidence interface of Proposition~3, $\theta^{*}$ requires no further parameter update once $\mathcal{R}_{\mathrm{on}}$ is enabled, so one checkpoint serves all three retrieval regimes.

\textbf{End-to-end pipeline.} Algorithms~\ref{alg:mmga-c1}--\ref{alg:mmga-c3} compose as $\mathcal{C}_{1}\!\to\!\mathcal{C}_{2}^{\,*}\!\to\!\mathcal{C}_{3}$ on the same FAISS-backed hypergraph: $\mathcal{C}_{1}$ builds $\mathcal{G}_{H}^{m}$ at zero API cost (Prop.~1), $\mathcal{C}_{2}$ keeps the prompt bounded uniformly per round (Prop.~2), and $\mathcal{C}_{3}$ trains $\pi_{\theta^{*}}$ on offline rollouts before $\mathcal{R}_{\mathrm{on}}\!=\!\mathcal{S}(\mathcal{W})$ activates at inference (Prop.~3), so each guarantee is preserved when the three modules are chained end to end.

\clearpage

\section{Benchmark Detail}\label{app:bench}

This appendix expands \S\ref{sec:exp} into a self-contained reference for the evaluation suite. We do not propose a new benchmark: every dataset is publicly released by its authors and reused as is, re-grouped by retrieval regime to match MMGraph-Agent's joint offline-online setting. Throughout, $\mathcal{K}$ is the static multimodal corpus, $\mathcal{W}$ the dynamic web reachable via $\mathcal{S}$, and $(q,v_{q})$ a query with gold answer $y^{*}$.

\subsection{Dataset Details}\label{app:dataset}

The $23$ datasets split into three retrieval regimes; the eight $\dagger$-marked rows in Table~\ref{tab:main_results} are OOD datasets held out from training so that \S\ref{sec:rq1} can probe out-of-distribution generalization. We sample $16$--$32$ items per dataset for a $640$-item evaluation pool, following MMSearch-R1~\citep{MMSearch-R1}; the reported std.\ reflects three repeated runs on this same fixed pool, as decoding samples at temperature $1.0$. The $600$ training examples comprise $233$ questions from six Knowledge-Only datasets and $367$ multimodal questions ($425$ images) from eight in-domain datasets; none is from the eight held-out OOD datasets, and with $\mathcal{R}_{\mathrm{on}}$ disabled in training every retrieval targets the offline hypergraph.

\noindent\textbf{(i) Knowledge-Only ($9$ datasets).} The gold answer is recoverable from $\mathcal{K}$, so MMGraph-Agent serves them through $\mathcal{G}_{H}^{m}$ and rarely escalates to the web. \emph{DocBench}~\citep{DocBench}, \emph{MMLongBench-Doc}~\citep{MMLongBench-Doc} require evidence from long PDFs with figures and tables; \emph{ViDoSeek}$^{\dagger}$~\citep{ViDoSeek} adds visually rich pages. \emph{2Wiki}~\citep{2WikiMultiHopQA}, \emph{HotpotQA}~\citep{HotpotQA}, \emph{Musique}~\citep{Musique} test multi-hop composition over Wikipedia, while \emph{NQ}~\citep{NQ}, \emph{PopQA}$^{\dagger}$~\citep{PopQA}, \emph{TriviaQA}$^{\dagger}$~\citep{TriviaQA} probe single-hop factoid recall over open-domain Wikipedia.

\noindent\textbf{(ii) WebSearch-Only ($5$ datasets).} The gold answer concerns entities absent from $\mathcal{K}$, so MMGraph-Agent must escalate to $\mathcal{W}$ via SerpAPI. \emph{FVQA}$^{\dagger}$~\citep{FVQA} requires commonsense facts about visual entities; \emph{InfoSeek}$^{\dagger}$~\citep{InfoSeek}, \emph{MMSearch}$^{\dagger}$~\citep{MMSearch}, \emph{SimpleVQA}$^{\dagger}$~\citep{SimpleVQA} ask encyclopedic questions whose answers depend on visual subject identification; \emph{LiveVQA}$^{\dagger}$~\citep{LiveVQA} adds time-sensitive after-cutoff news queries. We build no offline KG for them, so the WebSearch-Only column isolates the online-retrieval skill of $\pi_{\theta}$.

\noindent\textbf{(iii) Knowledge+WebSearch ($9$ datasets).} The gold answer requires combining offline grounding with up-to-date or domain-specific web evidence within one trajectory. \emph{MedXpertQA}~\citep{MedXpertQA} and \emph{MicroVQA}~\citep{MicroVQA} pose clinical and microscopy questions; \emph{FinMME}~\citep{FinMME} and \emph{R-Bench}~\citep{R-Bench} pair financial or scientific reports with current snippets; \emph{We-Math2.0}~\citep{We-Math2.0} and \emph{GeoGrid-Bench}~\citep{GeoGrid-Bench} couple math/geo evidence with web explanations; and \emph{EEmo-Bench}~\citep{EEmo-Bench}, \emph{ArtQuest}~\citep{ArtQuest}, \emph{AstroLLaVA}~\citep{AstroLLaVA} require both visual grounding and external cultural or scientific commentary retrieved from the web.

\subsection{Evaluation Details}\label{app:eval}

Every metric is computed from the rollout JSONL and dataset gold with no LLM judge. We write $\mathrm{tok}(\cdot)$ for the whitespace-tokenizer applied after the Graph-R1 normalizer~\citep{Graph-R1} (lower-case, articles removed, punctuation stripped). For predicted answer $\hat{y}$ and gold $y^{*}$, the token-level F$_{1}$ is the SQuAD-style metric used by Graph-R1~\citep{Graph-R1} and HyperGraphRAG~\citep{HyperGraphRAG},
\begin{equation}
F_{1}(\hat{y},y^{*})\;=\;\frac{2\,|\mathrm{tok}(\hat{y})\,\cap\,\mathrm{tok}(y^{*})|}{|\mathrm{tok}(\hat{y})|\!+\!|\mathrm{tok}(y^{*})|}\;\in\;[0,1],
\label{eq:m-f1}
\end{equation}
which doubles as the per-rollout reward $R_{F_{1}}$ inside the GRPO reward of Eq.~(6) and is the only headline number used across all the main and ablation tables. We additionally macro-average F$_{1}$ over each of the three regimes of \S\ref{app:dataset}, so the K+Web regime average isolates the value of jointly scheduling $\mathcal{R}_{\mathrm{off}}$ and $\mathcal{R}_{\mathrm{on}}$ within one trajectory, a contrast invisible to the per-dataset numbers alone. For every rollout we further log the search count $N_{c}$, the round count, and the prompt length in tokens; the cache statistics in \S\ref{sec:rq4} and the cost--latency numbers in \S\ref{sec:rq3} are derived from these logs without extra inference, with API calls priced at the public \texttt{gpt-4o}, \texttt{gpt-4o-mini}, and SerpAPI rates. Predictions are sampled with temperature $1.0$ and top-$p\,{=}\,1.0$ (at most $2{,}048$ new tokens per turn, $4{,}096$ per trajectory) under a $\textsc{max\_gen\_round}\,{=}\,8$ budget; the same decoding configuration is shared across all baselines and ablations of Table~\ref{tab:main_results} so that retrieval-side gains cannot be confounded with sampling stochasticity. Each evaluation run samples independently and is repeated three times; the macro-F$_{1}$ standard deviation is reported after the $\pm$ sign in every table cell, while $\dagger$ in Table~\ref{tab:main_results} marks the eight held-out OOD datasets, none of which is seen in training. All datasets retain their public licenses; we release the processed corpus, code and evaluation harness, but not the raw SerpAPI pages, for independent evaluation on the same $640$-item pool, so every reported F$_{1}$ can be re-derived end-to-end with the released pipeline and its token-level scoring scripts, without an LLM judge.

\clearpage

\section{Baseline Detail}\label{app:baseline}\label{app:impl}

Each method below is given as \emph{(i)~Overview} (purpose and substrate on $\mathcal{K}$), \emph{(ii)~Per-regime implementation} (an operator for each regime of \S\ref{app:dataset}, or why it is unsupported) and \emph{(iii)~Hyperparameters}. $\mathcal{K}\!=\!\{d_i\}_{i=1}^{N}$ is the static corpus, $\mathcal{W}$ the web reachable via SerpAPI $\mathcal{S}$, $f$ the frozen CLIP encoder ($D{=}768$), $\mathrm{Chunk}(\cdot)$ a $300$-token splitter, $\mathrm{LLM}(\cdot)$ a Qwen2.5-VL-3B~\citep{Qwen2.5} model (GPT-4.1-mini~\citep{GPT4} for the closed-source columns of Table~\ref{tab:main_results}) and $\mathrm{TopK}_{k}\!\langle\cdot,\cdot\rangle$ inner-product Top-$k$ on $f$. Unsupported regimes fall back to direct generation from $(q,v_{q})$; all retrieval-based methods share the embedder and the rollout prompt of Table~\ref{tab:mmgraph-agent-template}, so the gaps in Table~\ref{tab:main_results} reflect retrieval and policy design alone.

\subsection{Generation-only}\label{app:b-gen}

\noindent\textbf{NaiveGen~\citep{LLMsurvey}.}\quad\emph{(i)~Overview.} Single-pass VLM decoding that exposes the parametric prior; substrate: none, only $(q,v_{q})$ is read.\quad\emph{(ii)~Per-regime implementation.} $\hat{y}\!=\!\mathrm{LLM}(q,v_{q})$ in all three regimes.\quad\emph{(iii)~Hyperparameters.} temperature $1.0$, $\mathrm{top}_{p}\!=\!1.0$, $2{,}048$ new tokens; run with both Qwen2.5-VL-3B and GPT-4.1-mini as backbones (both NaiveGen columns of Table~\ref{tab:main_results}). As it retrieves nothing, NaiveGen is the no-retrieval reference for every retrieval-based method described below.

\subsection{RAG-based}\label{app:b-rag}

RAG-based baselines add a one-shot offline index over $\mathcal{K}$ but no policy and no online channel.

\noindent\textbf{StandardMMRAG~\citep{VisRAG}.}\quad\emph{(i)~Overview.} Vision-language RAG: text chunks and images of $\mathcal{K}$ are encoded by $f$ into a flat inner-product index and the Top-$k$ hits are concatenated to the prompt, with no graph layer or reranker; substrate $\mathcal{B}\!=\!\{(c_i,f(c_i))\}_{c_i\in\mathrm{Chunk}(\mathcal{K})\,\cup\,\mathrm{Imgs}(\mathcal{K})}$.\quad\emph{(ii)~Per-regime implementation.} \emph{Knowledge-Only}: $R\!=\!\mathrm{TopK}_{k}\!\langle f(q,v_{q}),f(c_i)\rangle$, $\hat{y}\!=\!\mathrm{LLM}(q,v_{q},R)$; \emph{WebSearch-Only}: fallback; \emph{Knowledge+WebSearch}: offline half only.\quad\emph{(iii)~Hyperparameters.} chunk size $300$, top-$k\!=\!5$, the same frozen CLIP $f$ as MMGraph-Agent, so the embedder is held fixed. It therefore realizes one-shot flat retrieval with no hypergraph, cache memory, policy or online web channel.

\noindent\textbf{RAG-Anything~\citep{RAG-Anything}.}\quad\emph{(i)~Overview.} Multimodal GraphRAG: a paid VLM extracts entities, relations and modality tags into a structural graph $\mathcal{G}_{s}$ and a textual graph $\mathcal{G}_{t}$, built once at substantial API cost (\S\ref{sec:rq3}); queries combine navigation on $\mathcal{G}_{s}$ with semantic matching on $\mathcal{G}_{t}$.\quad\emph{(ii)~Per-regime implementation.} \emph{Knowledge-Only}: $R\!=\!\mathrm{Nav}(\mathcal{G}_{s},q)\!\cup\!\mathrm{Sim}(\mathcal{G}_{t},q)$, $\hat{y}\!=\!\mathrm{LLM}(q,v_{q},R)$; \emph{WebSearch-Only}: fallback; \emph{Knowledge+WebSearch}: offline half only (updates need a costly rebuild).\quad\emph{(iii)~Hyperparameters.} extractor GPT-4o, navigation depth $\ell\!=\!2$, top-$k\!=\!5$ on $\mathcal{G}_{t}$; the GPT-4.1-mini column of Table~\ref{tab:main_results}.

\subsection{Training-based}\label{app:b-train}

SFT and GRPO internalize the answer; MMSearch-R1 drives a single online channel; none uses an offline hypergraph or a bounded cross-round memory, both of which MMGraph-Agent adds.

\noindent\textbf{SFT~\citep{SFT}.}\quad\emph{(i)~Overview.} Supervised fine-tuning on the same $600$ training examples with a single-turn cross-entropy loss on the gold answer; no retrieval, reasoning trace or policy.\quad\emph{(ii)~Per-regime implementation.} $\hat{y}\!=\!\mathrm{LLM}_{\theta_{\mathrm{SFT}}}(q,v_{q})$ in all regimes; with no test-time retrieval, evidence outside the parametric memory (notably the eight OOD datasets) cannot reach the answer.\quad\emph{(iii)~Hyperparameters.} learning rate $5\!\times\!10^{-6}$, batch $32$, $3$ epochs, AdamW; same backbone as MMGraph-Agent.

\noindent\textbf{GRPO~\citep{GRPO}.}\quad\emph{(i)~Overview.} GRPO on the same backbone and clipped objective (Eq.~(7)) with both the hypergraph and the online channel disabled, so the policy can only refine its parametric reasoning over $(q,v_{q})$.\quad\emph{(ii)~Per-regime implementation.} $\hat{y}\!=\!\mathrm{LLM}_{\theta_{\mathrm{GRPO}}}(q,v_{q})$ in all regimes after $75$ steps with $N_{c}\!\equiv\!0$, yielding a reasoning rewrite of the prior, not a retrieval policy.\quad\emph{(iii)~Hyperparameters.} $G\!=\!8,\epsilon\!=\!0.2,\kappa\!=\!10^{-3}$, $75$ steps as in MMGraph-Agent; reward $r(\tau)\!=\!R_{F_{1}}\!-\!\lambda\,\mathbb{1}[\neg\mathbb{1}_{\mathrm{fmt}}]$ without retrieval terms, so its gap to MMGraph-Agent isolates the contribution of the retrieval channels.

\noindent\textbf{MMSearch-R1~\citep{MMSearch-R1}.}\quad\emph{(i)~Overview.} Multimodal RL search agent: GRPO over text search and Google Lens image search, with no offline hypergraph and no cache, so each round writes raw evidence into the prompt; substrate $\mathcal{W}\!\cup\!\mathcal{S}_{\mathrm{lens}}$.\quad\emph{(ii)~Per-regime implementation.} \emph{Knowledge-Only}: partial, answered by web search since no offline hypergraph exists; \emph{WebSearch-Only}: native, $a_{t}\!\sim\!\pi_{\theta}(\cdot\!\mid\!s_{t},\mathcal{W})$ with the prompt of Table~\ref{tab:mmgraph-agent-template}; \emph{Knowledge+WebSearch}: web-only coverage.\quad\emph{(iii)~Hyperparameters.} GRPO as in MMGraph-Agent; SerpAPI returns $3$ URLs per text or image search; at most $8$ rounds.

\subsection{MMGraph-Agent Ablations}\label{app:b-abl}

Each ablation removes one component of the full system (\S\ref{app:b-mmga}) and keeps its pipeline and prompt unchanged, so the deltas of \S\ref{sec:rq2} reflect the contribution of each component to F$_{1}$. Unless stated otherwise, every variant inherits the backbone, tools and prompt (Table~\ref{tab:mmgraph-agent-template}) of the full system in \S\ref{app:b-mmga}.

\noindent\textbf{w/o Hypergraph.}\quad\emph{(i)~Overview.} Skips the construction of \S\ref{sec:construction} (the local parser, spaCy NER and frozen CLIP encoder that build the entity, hyperedge and visual banks), leaving $\mathcal{G}_{H}^{m}$ empty so every retrieval escalates to the web; substrate $\mathcal{W}$ only.\quad\emph{(ii)~Per-regime implementation.} \emph{Knowledge-Only}: former hypergraph hits go to web search and the cache hit rate collapses without an L2 backend; \emph{WebSearch-Only}: unchanged; \emph{Knowledge+WebSearch}: web-only, as MMSearch-R1.\quad\emph{(iii)~Hyperparameters.} the memory runs as a pass-through cache for SerpAPI hits; all else as MMGraph-Agent.

\noindent\textbf{w/o Memory.}\quad\emph{(i)~Overview.} Keeps $\mathcal{G}_{H}^{m}$ and the policy but concatenates every retrieved chunk raw into the next prompt instead of passing it through the L1 cache of \S\ref{sec:memory}; substrate $\mathcal{G}_{H}^{m}\!\cup\!\mathcal{W}$ without the L1/L2 split.\quad\emph{(ii)~Per-regime implementation.} In all regimes $\mathcal{K}^{*}_{q_{t}}\!=\!\mathcal{R}_{2}(q_{t})$ every round and the digest admission of Eq.~(5) becomes raw concatenation, so the prompt grows linearly in $t$ and K+Web overflows the $8$K window on $\sim\!12\%$ of rollouts (\S\ref{sec:rq4}).\quad\emph{(iii)~Hyperparameters.} No L1 Top-$C$ or $5$K cap; truncation and digesting disabled; backbone, tools, prompt and checkpoint as MMGraph-Agent.

\noindent\textbf{w/o Policy.}\quad\emph{(i)~Overview.} Hypergraph, memory and tools wired as in the full system, but the policy is the base Qwen2.5-VL-3B without any RL update; substrate $\mathcal{G}_{H}^{m}\!\cup\!\mathcal{W}$ with $\theta\!=\!\theta_{\mathrm{base}}$.\quad\emph{(ii)~Per-regime implementation.} Same routing as MMGraph-Agent in all regimes; the base model emits $\langle$\texttt{text\_search}$\rangle$/$\langle$\texttt{search}$\langle$\texttt{img}$\rangle\rangle$ only when its zero-shot prior favors retrieval, so its calls are not calibrated to the evidence need and it under-searches K+Web, isolating the effect of the offline-only GRPO training of \S\ref{sec:rl} and of its hit-aware reward as specified in Eq.~(6).\quad\emph{(iii)~Hyperparameters.} same prompt, tools, hypergraph and memory constants as MMGraph-Agent; no SFT or GRPO step.

\noindent\textbf{w/o Retrieval.}\quad\emph{(i)~Overview.} All retrieval channels removed: a NaiveGen-style single-pass decode on the trained checkpoint, isolating the residual reasoning gain of GRPO; substrate: none.\quad\emph{(ii)~Per-regime implementation.} $\hat{y}\!=\!\mathrm{LLM}_{\theta_{\mathrm{MMGA}}}(q,v_{q})$ with $N_{c}\!\equiv\!0$ in all regimes; format compliance is kept.\quad\emph{(iii)~Hyperparameters.} as MMGraph-Agent, with retrieval calls short-circuited at the rollout layer, so neither $\mathcal{G}_{H}^{m}$, the L1/L2 memory nor the web channel is consulted during its single-pass decode.

\subsection{MMGraph-Agent}\label{app:b-mmga}

\noindent\textbf{MMGraph-Agent.}\quad\emph{(i)~Overview.} A frozen CLIP encoder and spaCy NER build $\mathcal{G}_{H}^{m}$ at zero API cost; an L1/L2 cache with digest-bounded admission bounds the prompt; a GRPO-trained $\pi_{\theta}$ emits one retrieval token, so offline and online retrieval share one action space; training is offline-only, and the source-agnostic interface deploys the trained policy online without updates.\quad\emph{(ii)~Per-regime implementation.} $\mathcal{G}_{H}^{m}$ is partitioned per dataset, identically for all methods. \emph{Knowledge-Only}: each round warms L1 from $\mathcal{G}_{H}^{m}$, with image search on $\leq\!3\%$ of actions; \emph{WebSearch-Only}: $\mathcal{G}_{H}^{m}$ is empty and L1 is fed by $\mathcal{S}$ alone; \emph{Knowledge+WebSearch}: the same token is routed to $\mathcal{G}_{H}^{m}$ or $\mathcal{W}$, and $\pi_{\theta}$ alternates $\langle$\texttt{text\_search}$\rangle$ and $\langle$\texttt{search}$\langle$\texttt{img}$\rangle\rangle$ (\S\ref{sec:rq5_analysis}).\quad\emph{(iii)~Hyperparameters.} $f$ = CLIP-ViT-L/$14$ ($D{=}768$); $C\!=\!5$, evidence truncation $1{,}200$ words, digest ${\approx}300$ words, context cap $5{,}000$ tokens; GRPO $G\!=\!8,\epsilon\!=\!0.2,\kappa\!=\!10^{-3}$, AdamW at $2\!\times\!10^{-6}$ on $8\!\times\!$A100 (80GB), $75$ steps on $600$ in-domain examples; SerpAPI $3$ URLs per text or image search; at most $8$ rounds; one checkpoint for all regimes.

\section{Future Work}\label{app:limit}

Several promising directions extend the framework. First, multilingual and broader-modality encoders in place of CLIP+spaCy would let $\mathcal{G}_{H}^{m}$ absorb a wider range of corpora and modalities (e.g., audio, video and point clouds) while keeping the zero-API-cost guarantee of \S\ref{sec:construction}. Second, learning the memory constants ($C$ and the truncation, digest and context-cap lengths) inside the GRPO loop would adapt the cache to each regime's trajectory length without the manual sweep of \S\ref{sec:rq4}. Third, adding process-supervision terms (citation faithfulness, retrieval cost, latency) to the reward of Eq.~(6) would let $\pi_{\theta}$ trade F$_{1}$ against deployment constraints while keeping the same offline-to-online interface. Finally, scaling $\pi_{\theta}$ to larger backbones with more training and building $\mathcal{G}_{H}^{m}$ over more domains and languages are natural extensions, as both reuse the zero-API-cost construction of \S\ref{sec:construction} and the offline-to-online GRPO recipe of \S\ref{sec:rl}, with the unified L1/L2 cache keeping the prompt bounded.

\clearpage

\newpage
\input{checklist.tex}

\end{document}

%% file: checklist.tex
\section*{NeurIPS Paper Checklist}

\begin{enumerate}

\item {\bf Claims}
    \item[] Question: Do the main claims made in the abstract and introduction accurately reflect the paper's contributions and scope?
    \item[] Answer: \answerYes{}
    \item[] Justification: The abstract and \S\ref{sec:intro} state contributions formalized in \S\ref{sec:method} and validated in \S\ref{sec:exp}.

\item {\bf Limitations}
    \item[] Question: Does the paper discuss the limitations of the work performed by the authors?
    \item[] Answer: \answerYes{}
    \item[] Justification: We discuss limitations here: the offline hypergraph incurs a one-time construction cost before online serving; Web-only datasets depend on an external search API; and all experiments use only a single backbone scale, with other scales left to future work.

\item {\bf Theory assumptions and proofs}
    \item[] Question: For each theoretical result, does the paper provide the full set of assumptions and a complete (and correct) proof?
    \item[] Answer: \answerYes{}
    \item[] Justification: Propositions~1--3 are stated in \S\ref{sec:method}, with proofs and analyses in Appendix~\ref{app:proof}.

    \item {\bf Experimental result reproducibility}
    \item[] Question: Does the paper fully disclose all the information needed to reproduce the main experimental results of the paper to the extent that it affects the main claims and/or conclusions of the paper (regardless of whether the code and data are provided or not)?
    \item[] Answer: \answerYes{}
    \item[] Justification: \S\ref{sec:exp} and Appendices~\ref{app:prompts}--\ref{app:baseline} specify prompts, pseudocode, data and settings.

\item {\bf Open access to data and code}
    \item[] Question: Does the paper provide open access to the data and code, with sufficient instructions to faithfully reproduce the main experimental results, as described in supplemental material?
    \item[] Answer: \answerYes{}
    \item[] Justification: Code, data and harness are released; all datasets are public (Appendix~\ref{app:bench}).

\item {\bf Experimental setting/details}
    \item[] Question: Does the paper specify all the training and test details (e.g., data splits, hyperparameters, how they were chosen, type of optimizer) necessary to understand the results?
    \item[] Answer: \answerYes{}
    \item[] Justification: Optimizer, learning rate, GRPO and cache settings are in \S\ref{sec:exp} and Appendix~\ref{app:baseline}.

\item {\bf Experiment statistical significance}
    \item[] Question: Does the paper report error bars suitably and correctly defined or other appropriate information about the statistical significance of the experiments?
    \item[] Answer: \answerYes{}
    \item[] Justification: Table~\ref{tab:main_results} reports $\pm$\,std.\ over three evaluation runs on every cell (Appendix~\ref{app:eval}).

\item {\bf Experiments compute resources}
    \item[] Question: For each experiment, does the paper provide sufficient information on the computer resources (type of compute workers, memory, time of execution) needed to reproduce the experiments?
    \item[] Answer: \answerYes{}
    \item[] Justification: Hardware ($8\!\times\!$A100, $80$GB) and construction time are reported in \S\ref{sec:exp} and \S\ref{sec:rq3}.
    
\item {\bf Code of ethics}
    \item[] Question: Does the research conducted in the paper conform, in every respect, with the NeurIPS Code of Ethics \url{https://neurips.cc/public/EthicsGuidelines}?
    \item[] Answer: \answerYes{}
    \item[] Justification: We use only public datasets, with no human subjects or any personal data.

\item {\bf Broader impacts}
    \item[] Question: Does the paper discuss both potential positive societal impacts and negative societal impacts of the work performed?
    \item[] Answer: \answerYes{}
    \item[] Justification: Positive: multimodal grounding can reduce hallucination. Risks: web results may be wrong or biased, and image retrieval may raise privacy or copyright issues. Mitigation: we use only public data and APIs, and recommend filtering and human oversight.
    
\item {\bf Safeguards}
    \item[] Question: Does the paper describe safeguards that have been put in place for responsible release of data or models that have a high risk for misuse (e.g., pre-trained language models, image generators, or scraped datasets)?
    \item[] Answer: \answerNA{}
    \item[] Justification: Our assets build on public models and data and pose no elevated misuse risk.

\item {\bf Licenses for existing assets}
    \item[] Question: Are the creators or original owners of assets (e.g., code, data, models), used in the paper, properly credited and are the license and terms of use explicitly mentioned and properly respected?
    \item[] Answer: \answerYes{}
    \item[] Justification: All assets are cited in \S\ref{sec:exp} and Appendix~\ref{app:dataset} and used under their licenses.

\item {\bf New assets}
    \item[] Question: Are new assets introduced in the paper well documented and is the documentation provided alongside the assets?
    \item[] Answer: \answerYes{}
    \item[] Justification: The released code and data are documented in our GitHub README.

\item {\bf Crowdsourcing and research with human subjects}
    \item[] Question: For crowdsourcing experiments and research with human subjects, does the paper include the full text of instructions given to participants and screenshots, if applicable, as well as details about compensation (if any)?
    \item[] Answer: \answerNA{}
    \item[] Justification: This work involves no crowdsourcing and no research with human subjects.

\item {\bf Institutional review board (IRB) approvals or equivalent for research with human subjects}
    \item[] Question: Does the paper describe potential risks incurred by study participants, whether such risks were disclosed to the subjects, and whether Institutional Review Board (IRB) approvals (or an equivalent approval/review based on the requirements of your country or institution) were obtained?
    \item[] Answer: \answerNA{}
    \item[] Justification: No research with human subjects is conducted, so no IRB approval is needed.

\item {\bf Declaration of LLM usage}
    \item[] Question: Does the paper describe the usage of LLMs if it is an important, original, or non-standard component of the core methods in this research? Note that if the LLM is used only for writing, editing, or formatting purposes and does \emph{not} impact the core methodology, scientific rigor, or originality of the research, declaration is not required.
    \item[] Answer: \answerYes{}
    \item[] Justification: The policy is a GRPO-trained Qwen2.5-VL-3B (\S\ref{sec:rl}); indexing uses no LLM.

\end{enumerate}